\documentclass{article}

\usepackage{color}
\usepackage[table]{xcolor}
\usepackage{multirow}
\usepackage{adjustbox}

\usepackage{CJKutf8}

\newcommand{\tb}[1]{\textbf{#1}}
\newcommand{\ud}[1]{\underline{#1}}

\usepackage{microtype}
\usepackage{graphicx}
\usepackage{subcaption}
\usepackage{booktabs} % for professional tables

\usepackage{hyperref}

\usepackage[accepted]{icml2026}

\usepackage{amsmath}
\usepackage{amssymb}
\usepackage{mathtools}
\usepackage{amsthm}
\usepackage{comment}

\usepackage[capitalize,noabbrev]{cleveref}

\theoremstyle{plain}

\theoremstyle{definition}

\theoremstyle{remark}

\usepackage[textsize=tiny]{todonotes}

\icmltitlerunning{ML-Embed: Inclusive and Efficient Embeddings for a Multilingual World}

\begin{document}

\twocolumn[
  \icmltitle{ML-Embed: Inclusive and Efficient Embeddings for a Multilingual World}



  \icmlsetsymbol{equal}{*}

  \begin{icmlauthorlist}
    \icmlauthor{Ziyin Zhang}{sjtu,ant}
    \icmlauthor{Zihan Liao}{ant}
    \icmlauthor{Hang Yu}{ant}
    \icmlauthor{Peng Di}{ant}
    \icmlauthor{Rui Wang}{sjtu}
  \end{icmlauthorlist}

  \icmlaffiliation{sjtu}{School of Computer Science, Shanghai Jiao Tong University, Shanghai, China}
  \icmlaffiliation{ant}{Ant Group, Hangzhou, China}

  \icmlcorrespondingauthor{Hang Yu}{hyu.hugo@antgroup.com}
  \icmlcorrespondingauthor{Peng Di}{dipeng.dp@antgroup.com}
  \icmlcorrespondingauthor{Rui Wang}{wangrui12@sjtu.edu.cn}

  \icmlkeywords{Machine Learning, ICML}

  \vskip 0.3in
]



\printAffiliationsAndNotice{}  % no special notice (required even if empty)

\begin{abstract}
  % The development of high-quality text embeddings is increasingly drifting toward an exclusionary future, defined by three critical barriers: prohibitive computational costs, a narrow linguistic focus that neglects most of the world's languages, and a lack of transparency from closed-source or open-weight models that stifles research. To dismantle these barriers, we introduce ML-Embed, a suite of inclusive and efficient models built upon a new framework: 3-Dimensional Matryoshka Learning (3D-ML). Our framework addresses the computational challenge with comprehensive efficiency across the entire model lifecycle. Beyond the storage benefits of Matryoshka Representation Learning (MRL), we integrate Matryoshka Layer Learning (MLL) for flexible inference-time depth and Matryoshka Embedding Learning (MEL) for enhanced parameter efficiency. To address the linguistic challenge, we curate a massively multilingual dataset and train a suite of models ranging from 140M to 8B parameters. In a direct commitment to transparency, we release all models, data, and code. Extensive evaluation on 430 tasks demonstrates that our models set new records on 9 of 17 evaluated MTEB benchmarks, with particularly strong results in low-resource languages, providing a reproducible blueprint for building globally equitable and computationally efficient AI systems.
  The development of high-quality text embeddings is increasingly drifting toward an exclusionary future, defined by three critical barriers: prohibitive computational costs, a narrow linguistic focus that neglects most of the world's languages, and a lack of transparency from closed-source or open-weight models that stifles research. To dismantle these barriers, we introduce ML-Embed, a suite of inclusive and efficient models built upon a new framework: 3-Dimensional Matryoshka Learning (3D-ML). Our framework addresses the computational challenge with comprehensive efficiency across the entire model lifecycle. Beyond the storage benefits of Matryoshka Representation Learning (MRL) and flexible inference-time depth provided by Matryoshka Layer Learning (MLL), we introduce Matryoshka Embedding Learning (MEL) for enhanced parameter efficiency. To address the linguistic challenge, we curate a massively multilingual dataset and train a suite of models ranging from 140M to 8B parameters. In a direct commitment to transparency, we release all models, data, and code. Extensive evaluation on 430 tasks demonstrates that our models set new records on 9 of 17 evaluated MTEB benchmarks, with particularly strong results in low-resource languages, providing a reproducible blueprint for building globally equitable and computationally efficient AI systems.
\end{abstract}

\section{Introduction}\label{sec:introduction}

Text embeddings are a foundational component of modern AI, translating the richness of human language into numerical representations that dictate the performance, fairness, and accessibility of the downstream systems they enable, from semantic search to Retrieval-Augmented Generation (RAG)~\citep{2023RAG-Survey}.

However, the paradigm for developing state-of-the-art embedding models has shifted toward repurposing massive decoder-based language models. While powerful, this trend is creating a critical \textbf{computational barrier} characterized by prohibitive training costs and immense memory footprints. This computational barrier exacerbates a growing linguistic barrier: as models become more resource-intensive, they become increasingly inaccessible to the broader research community and undeployable in resource-constrained environments where many of the world's low-resource languages are spoken. While techniques like Matryoshka Representation Learning (MRL, \citealp{2022MRL}) offer partial relief by optimizing storage, they leave the immense burdens of training and inference untouched.

% To dismantle this computational barrier, we introduce \textbf{3-Dimensional Matryoshka Learning (3D-ML)}, a unified framework that generalizes the nested learning principle to ensure comprehensive efficiency across the entire model lifecycle. 3D-ML integrates three synergistic axes: \textbf{Matryoshka Embedding Learning (MEL) for parameter efficiency}, \textbf{Matryoshka Layer Learning (MLL)} for adaptable inference depth, and the established \textbf{Matryoshka Representation Learning (MRL)} for storage efficiency. Crucially, these components do not merely operate in isolation; they function synergistically to allow for deeper, more capable architectures that fit within strict parameter and compute budgets, creating a holistic solution to the efficiency problem.

To dismantle this computational barrier, we introduce 3-Dimensional Matryoshka Learning (3D-ML), a unified framework built upon Matryoshka Layer Learning (MLL, \citealp{2024MLL}) and Matryoshka Representation Learning (MRL, \citealp{2022MRL}) while integrating a novel Matryoshka Embedding Learning (MEL) technique that addresses the critical challenge of parameter-heavy embedding layers by learning two factorized, low-rank matrices that are themselves structured for nested training. This provides significant parameter savings for both training and inference and offers flexible deployment options that balance efficiency with compatibility.

To validate the practical utility of 3D-ML, we applied it to the notoriously resource-intensive challenge of creating massively multilingual models—a domain that faces two further critical gaps. The first is a \textbf{linguistic challenge}: despite comprehensive benchmarks like MTEB~\citep{2023MTEB,2025MMTEB}, research attention remains disproportionately focused on a few high-resource languages. As illustrated in Table~1, submissions to benchmarks for languages like Polish, Persian, and Vietnamese are orders of magnitude fewer than for English. The second is a \textbf{transparency challenge}: progress is stymied by a lack of openness, as many top-performing models~\citep{2025Qwen3-Embedding, 2025Gemini-Embedding} are released as closed-source APIs or as open-weight models with no training transparency, hindering reproducible research. Addressing these interconnected issues, we introduce ML-Embed, a family of efficient and inclusive models built with our 3D-ML framework and a new, massively multilingual dataset.

This work makes the following contributions:
\begin{itemize}
    % \item We propose 3-Dimensional Matryoshka Learning (3D-ML), a novel framework that integrates three nested learning techniques (MEL, MLL, and MRL) to provide end-to-end efficiency for the training, inference, and storage of embedding models.
    \item We propose MEL, a novel efficient training technique. Integrating MEL with MLL and MRL, we present a 3-Dimensional Matryoshka Learning framework, providing end-to-end efficiency for the training, inference, and storage of embedding models.
    \item We demonstrate that efficiency and inclusivity can drive superior performance. ML-Embed-8B establishes new state-of-the-art results on 9 of 17 MTEB benchmarks. Crucially, we achieve massive gains in historically underserved languages—such as a +22.89 point improvement on Polish and +6.88 on Vietnamese—proving that equitable performance need not come at the cost of efficiency.
    \item In a direct counter to the trend of closed development, we release our comprehensive multilingual dataset, all model weights, and training code, providing a fully reproducible blueprint for building globally equitable AI systems\footnote{Code: \url{https://github.com/codefuse-ai/CodeFuse-Embeddings}. Model and data: \url{https://huggingface.co/collections/codefuse-ai/codefuse-embeddings}.}.
\end{itemize}

\section{Related Work}

\subsection{Efficient Representation Learning}

Matryoshka Representation Learning~\citep{2022MRL} optimizes $d$-dimensional embeddings by applying loss functions at $O(\log (d))$ embedding sizes, facilitating adaptive application in downstream tasks with varying dimension requirements. Recent extensions such as ESE~\citep{2025ESE} improve MRL by applying principal component analysis to condense more essential information into the initial embedding dimensions and model layers, while Matryoshka-Adaptor~\citep{2024Matryoshka-adapter} and SMEC~\citep{2025SMEC} employ additional MLP layers to reduce the embeddings to lower dimensions. Other methods, such as Flextron~\citep{2024Flextron} and MatFormer~\citep{2024MetaFormer}, also enable flexible model sizes by pruning attention heads or MLP dimensions at inference time. However, these methods often introduce structural modifications (e.g., routing mechanisms) that may reduce compatibility and complicate deployment.

In terms of training, existing matryoshka optimization methods focus on representation flexibility but do not reduce trainable parameters or the overall training cost, making them ill-suited to data-constrained or compute-constrained training scenarios. In these settings, parameter-efficient finetuning methods such as LoRA~\citep{2022LoRA} are used instead, which reduces the model's trainable parameters by decomposing the \emph{update} in weight matrices into low-rank ones. Numerous variants of LoRA have also been proposed, such as QLoRA~\citep{2023QLoRA} that combines LoRA and quantization, AdaLoRA~\citep{2023AdaLoRA} that adaptively allocates the parameter budget according to the importance of weight matrices, and RaSA~\citep{2025RaSA} that partially shares LoRA parameters across model layers. Nevertheless, all these methods require the entire model to be loaded at inference time, limiting their utility for resource-constrained deployment where reduced memory footprints are required.

\begin{table}
    \centering
    \caption{Number of models with complete results on MTEB benchmarks. While Multilingual and English have become popular testbeds for embedding models, some languages - especially Polish, Japanese, Vietnamese, and Persian - receive far less attention.}
    \label{tab:num-submission}
    \small
    \begin{tabular}{lr}
    \toprule
        \tb{Benchmark} & \tb{Models} \\
    \midrule
        Multilingual & 146 \\
        English & 154 \\
        European & 122 \\
        Indic & 111 \\
        Scandinavian & 41 \\
        Chinese & 43 \\
        German & 90 \\
        French & 111 \\
        Japanese & 11 \\
        Korean & 95 \\
        Dutch & 87 \\
        Polish & 1 \\
        Russian & 95 \\
        Persian & 22 \\
        Vietnamese & 17 \\
    \bottomrule
    \end{tabular}
\end{table}

\subsection{Multilingual Embedding Models and Benchmarks}

The previous generation of encoder-based embedding models witnessed a proliferation of massively multilingual embedding models supporting hundreds of languages, represented by XLM-R~\citep{2020XLM-R}, mDeBERTaV3~\citep{2023DebertaV3}, mBART~\citep{2020mBART}, and mT5~\citep{2021mT5}. Recently, decoder-based embedding models have become the dominant paradigm, benefiting from their extensive capabilities acquired during large-scale pre-training, as verified by state-of-the-art models such as E5-Mistral~\citep{2024E5-Mistral}, NV-Embed~\citep{2025NV-Embed}, Qwen3-Embedding~\citep{2025Qwen3-Embedding}, and Gemini-Embedding~\citep{2025Gemini-Embedding}.

However, this advancement has been accompanied by a shift toward English-centric evaluation. This is evidenced in MTEB~\citep{2023MTEB}, which has been established as one of the most recognized text embedding benchmarks, covering over 500 evaluation tasks and more than 250 languages~\citep{2025MMTEB}. Yet, in reality, the MTEB leaderboards exhibit significant linguistic bias. For instance, in the MTEB-Multilingual benchmark, 35 out of the 131 tasks focus exclusively on English, potentially obscuring a model's true multilingual efficacy. Furthermore, as illustrated in Table~\ref{tab:num-submission}, many language-specific benchmarks receive disproportionately less attention compared with the English or Multilingual benchmarks\footnote{All references to MTEB leaderboards in this manuscript refer to the snapshot acquired on January 22nd, 2026.}.

This disparity is exacerbated by the fact that many top-performing multilingual embedding models - such as Qwen3-Embedding~\citep{2025Qwen3-Embedding}, Gemini-Embedding~\citep{2025Gemini-Embedding}, and EmbeddingGemma~\citep{2025EmbeddingGemma} - are either closed-source APIs or open-weight only without training transparency. KaLM-Embedding~\citep{2025KaLM-Embedding-V2} represents one of the few exceptions with transparency in training data, but focuses exclusively on the Multilingual leaderboard and is not evaluated on the aforementioned language-specific benchmarks that are critical for truly global applications.

\section{Method: 3D Matryoshka Learning}

\begin{figure}[t]
  \centering
  \includegraphics[width=0.9\linewidth]{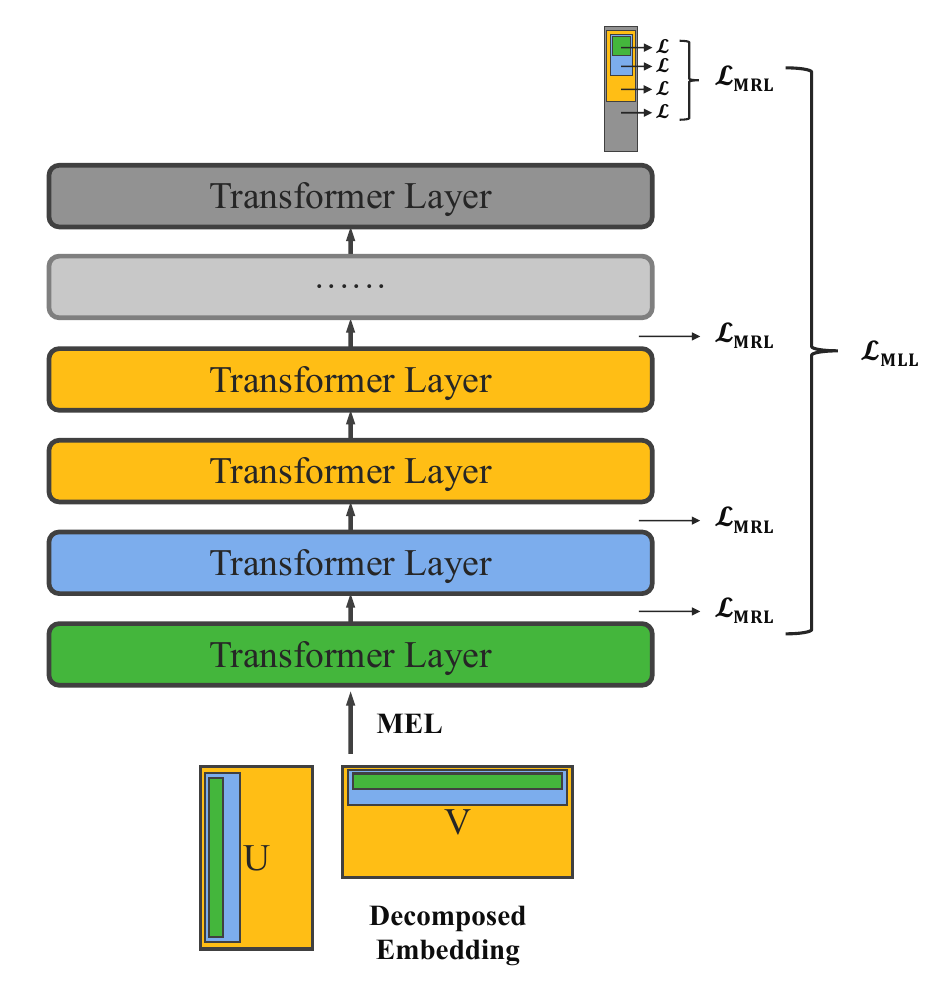}
  \caption{The 3D-ML framework provides comprehensive efficiency by applying nested learning principles across model parameters (MEL), depth (MLL), and representation dimensions (MRL).}
  \label{fig:3d-ml-overview}
\end{figure}

Creating truly accessible and scalable embedding models requires tackling efficiency bottlenecks across the entire model lifecycle: from the high costs of \textbf{training}, to the computational demands of \textbf{inference}, and finally to the footprint of \textbf{storage}. To this end, we propose \textbf{3-Dimensional Matryoshka Learning (3D-ML)}, a unified framework that generalizes the principle of nested structures to provide comprehensive efficiency. 3D-ML simultaneously targets all three stages by optimizing along three corresponding axes: model parameters, computational depth, and representation size. This is achieved through a trio of integrated techniques: 1) \textbf{Matryoshka Embedding Learning (MEL)} reduces trainable and total parameters for efficient training and inference; 2) \textbf{Matryoshka Layer Learning (MLL)} enables flexible model depth for efficient inference; 3) \textbf{Matryoshka Representation Learning (MRL)} produces variable-size representation dimensions for efficient storage. Figure~\ref{fig:3d-ml-overview} provides a conceptual illustration of this framework.

\subsection{Matryoshka Embedding Learning (MEL) for Parameter Efficiency}\label{sec:method-mel}

The embedding layer, which maps vocabulary tokens to dense vectors, often constitutes a disproportionately large share of a model's parameters, especially in smaller models and multilingual models with a large vocabulary. For instance, in an embedding model trained from Qwen3-0.6B~\citep{2025Qwen3}, the embedding layer accounts for 1/4 of the total parameters. MEL addresses this by learning the embedding matrix in a factorized, low-rank form that is itself structured for nested training. Crucially, unlike low-rank update methods such as LoRA~\citep{2022LoRA}, MEL reduces not only \emph{trainable parameters}, but also \emph{total parameters} so that inference efficiency is also improved.

Let the base model's original embedding matrix be $E \in \mathbb{R}^{v \times d_{\text{model}}}$, where $v$ is the vocabulary size and $d_{\text{model}}$ is the model's hidden dimension. Prior to fine-tuning, we initialize two smaller matrices, $E_A$ and $E_B$, using a truncated Singular Value Decomposition (SVD) of $E$. We compute $U, S, V^T = \text{SVD}(E)$ and select the top-$r$ singular values and vectors to form $U_r \in \mathbb{R}^{v \times r}$, $S_r \in \mathbb{R}^{r \times r}$, and $V_r^T \in \mathbb{R}^{r \times d_{\text{model}}}$. The trainable matrices are then initialized as:
\begin{equation}
    E_A \leftarrow U_r S_r \in \mathbb{R}^{v \times r} \quad \text{and} \quad E_B \leftarrow V_r^T \in \mathbb{R}^{r \times d_{\text{model}}}.
\end{equation}
The full embedding matrix is approximated by their product, $E \approx E_A E_B$. During fine-tuning, only $E_A$ and $E_B$ are updated instead of a full $v \times d_{\text{model}}$ matrix, reducing trainable parameters and memory requirements.

To embed the Matryoshka principle, during each training forward pass, we dynamically sample a sub-rank $r' < r$ from a predefined set (e.g., $\{64, 128, 256, 512, 1024\}$). The forward pass then uses only the first $r'$ components of the factorized matrices:
\begin{equation}
    E_{\text{effective}} = E_A[:, :r'] E_B[:r', :].
\end{equation}
This forces the model to prioritize the most critical information within the initial dimensions of the factorized space.

At inference time, MEL offers two modes:
\begin{itemize}
    \item \tb{Compatibility Mode}: We compute the final trained embedding matrix $E_{\text{trained}} = E_A E_B$. This results in a standard embedding layer, requiring no changes to existing inference infrastructure while still benefiting from the regularized training via low-rank factorization.
    % \item \tb{Efficiency Mode}: For maximum resource savings, we can deploy a model with a rank even smaller than $r$. We first compute $E_{\text{trained}} = E_A E_B$, then perform SVD on this trained matrix: $U', S', {V'}^T = \text{SVD}(E_{\text{trained}})$. A highly compressed model can be deployed using the top $r' \ll r$ components, by storing only the two matrices $U'_{r'} S'_{r'}$ and ${V'_{r'}}^T$, leading to substantial reductions in model size with negligible performance loss.
    \item \tb{Efficiency Mode}: For maximum resource savings, we can deploy the model with a highly compressed embedding layer. After training, we can either use the trained factorized matrices $E_A$ and $E_B$ directly (at rank $r$) or re-factorize the full matrix $E_{\text{trained}} = E_A E_B$ to an even smaller rank $r' \ll r$ for aggressive compression. Let the new factorized matrices be $E'_A \in \mathbb{R}^{v \times r'}$ and $E'_B \in \mathbb{R}^{r' \times d_{\text{model}}}$. This approach yields two key benefits. First, it drastically reduces the storage space: instead of storing a dense $v \times d_{\text{model}}$ matrix, we only store $v \times r' + r' \times d_{\text{model}}$ parameters. For a large vocabulary $V$ and a small rank $r'$, this represents a substantial reduction. Second, it can improve computational efficiency. A standard embedding lookup for a sequence of tokens involves gathering rows from the large $v \times d_{\text{model}}$ matrix. With factorization, this becomes a two-step process: a fast lookup in the ``tall-and-skinny'' matrix $E'_A$ followed by a matrix multiplication with the ``short-and-wide'' matrix $E'_B$. This is particularly advantageous for on-device deployment where memory is the primary constraint.
\end{itemize}

\subsection{Matryoshka Layer Learning (MLL) for Inference Efficiency}

The computational cost of Transformer-based models scales with model depth. MLL is designed to produce models that can be dynamically and efficiently truncated to shallower depths without significant performance degradation.

Instead of applying the training loss only at the final layer's output, MLL applies it at multiple, pre-defined intermediate layers. Let $\mathcal{L}_{\text{layers}} = \{l_1, l_2, \dots, l_k, L\}$ be a set of selected layer indices, where $L$ is the index of the final layer. For our experiments, we use a logarithmically spaced set of layers (e.g., $\{1, 2, 4, 8, 16, 32\}$) plus the model's final layer. For each layer $l \in \mathcal{L}_{\text{layers}}$, we extract its hidden state output, $h_l$. To maintain representational consistency across depths, we pass each $h_l$ through the model's final layer normalization, $\text{LN}_{\text{final}}$, before using it to compute the loss.

This ``early-exit'' style training ensures that shallower versions of the model are also effective embedders. At inference time, this provides unparalleled flexibility: one can deploy a smaller, faster model by simply taking the first $l$ layers of the full model, where $l \in \mathcal{L}_{\text{layers}}$. This avoids the need for re-training or complex pruning, enabling seamless adaptation to varying computational budgets.

\subsection{Unifying the Framework with Matryoshka Representation Learning (MRL)}

The final dimension of our framework is Matryoshka Representation Learning (MRL; \citealp{2022MRL}), which optimizes embeddings for variable-dimension storage. MRL trains a model such that prefixes of the final embedding vector are themselves effective, lower-dimensional representations.

In 3D-ML, MRL is not a separate step but is deeply integrated with MLL. For each selected MLL layer $l \in \mathcal{L}_{\text{layers}}$, we apply a contrastive loss not just on the full-dimensional output representation, but on a nested set of its prefixes. Let $\mathcal{D}_{\text{mrl}}$ be the set of MRL dimensions (e.g., $\{8, 16, 32, \dots, d_{\text{model}}\}$). Let $\text{proj}_d(v)$ denote the projection of a vector $v$ to its first $d$ dimensions. The total loss for a given layer $l$ is a sum of losses over these dimensions.

The unified 3D-ML objective function combines all three components. The total loss is summed over all selected MLL layers and all MRL dimensions. Let $h_l(q)$ and $h_l(d)$ be the hidden states from layer $l$ for a query $q$ and document $d$, respectively. The final representation for a given MRL dimension $d'$ is $v_{l, d'}(\cdot) = \text{proj}_{d'}(\text{LN}_{\text{final}}(h_l(\cdot)))$. The overall objective is:
\begin{equation}
    \mathcal{L}_{\text{3D-ML}} = \sum_{l \in \mathcal{L}_{\text{layers}}} \sum_{d' \in \mathcal{D}_{\text{mrl}}} c_{l,d'} \mathcal{L}_{\text{cl}}(q_i, d_i^+, \{d_{i,j}^-\}_{j=1}^n; v_{l, d'}),
    \label{eq:3dml_loss}
\end{equation}
where $c_{l,d'}$ is the loss weight coefficient for layer $l$ and dimension $d'$, and the contrastive learning loss $\mathcal{L}_{\text{cl}}$ for a given representation function $v_{l, d'}$ is defined as:
\begin{equation}
    - \log\frac{e^{s(v_{l, d'}(q_i), v_{l, d'}(d_i^+))/\tau}}{e^{s(v_{l, d'}(q_i), v_{l, d'}(d_i^+))/\tau}+\sum\limits_{j=1}^{n}e^{s(v_{l, d'}(q_i), v_{l, d'}(d_{i,j}^-))/\tau}}.
    \label{eq:contrastive_loss}
\end{equation}
Here, $s(\cdot, \cdot)$ is cosine similarity, $\tau$ is a temperature hyperparameter, $d_i^+$ is a positive document for query $q_i$, and $\{d_{i,j}^-\}$ are hard negative documents for query $q_i$. This unified loss ensures that the model learns representations that are simultaneously efficient in terms of parameters (via MEL), depth (via MLL), and storage (via MRL).

\subsection{Practical Deployment and Compatibility}

A core design principle of the 3D-ML framework is its focus on practical deployment and compatibility with existing ecosystems, ensuring that its efficiency gains are not merely theoretical but easily accessible to practitioners. Each component is designed for minimal friction:

\begin{itemize}
    \item \textbf{MRL (Storage)}: The benefits of Matryoshka Representation Learning are straightforward to leverage. Truncating the final embedding vectors is a simple post-processing step that is natively supported in popular libraries like \textsc{sentence-transformers}~\citep{2019SBERT} via a single parameter, requiring no changes to the codebase.
    
    \item \textbf{MLL (Inference)}: Matryoshka Layer Learning is similarly user-friendly and fully compatible with the Hugging Face ecosystem. Deploying a faster, shallower model is as simple as modifying the \texttt{num\_hidden\_layers} parameter in the model's configuration file, which drives the AutoModel class from \textsc{transformers}~\citep{2020transformers} to load only the first $n$ layers and ignore the remaining weights.
    
    \item \textbf{MEL (Parameters)}: As described in Section~\ref{sec:method-mel}, Matryoshka Embedding Learning offers a flexible trade-off between convenience and maximum efficiency. In \textbf{compatibility mode}, the factorized matrices ($E_A, E_B$) are multiplied into a standard embedding matrix before release, making the model indistinguishable from a standard Transformer decoder, ensuring seamless integration into any inference pipeline without code changes. For users prioritizing a minimal memory footprint, the \textbf{efficiency mode} allows for deploying the low-rank factorized matrices directly, enabling significant parameter reduction with only minor adjustments to the modeling file.
\end{itemize}

This comprehensive focus on deployability makes 3D-ML a practical blueprint for building and sharing highly efficient models, lowering the barrier to entry for a wide range of users, from large-scale production systems to resource-constrained research environments.

\section{Training Data}\label{sec:data}

\begin{figure*}
    \centering
    \includegraphics[width=1.0\linewidth]{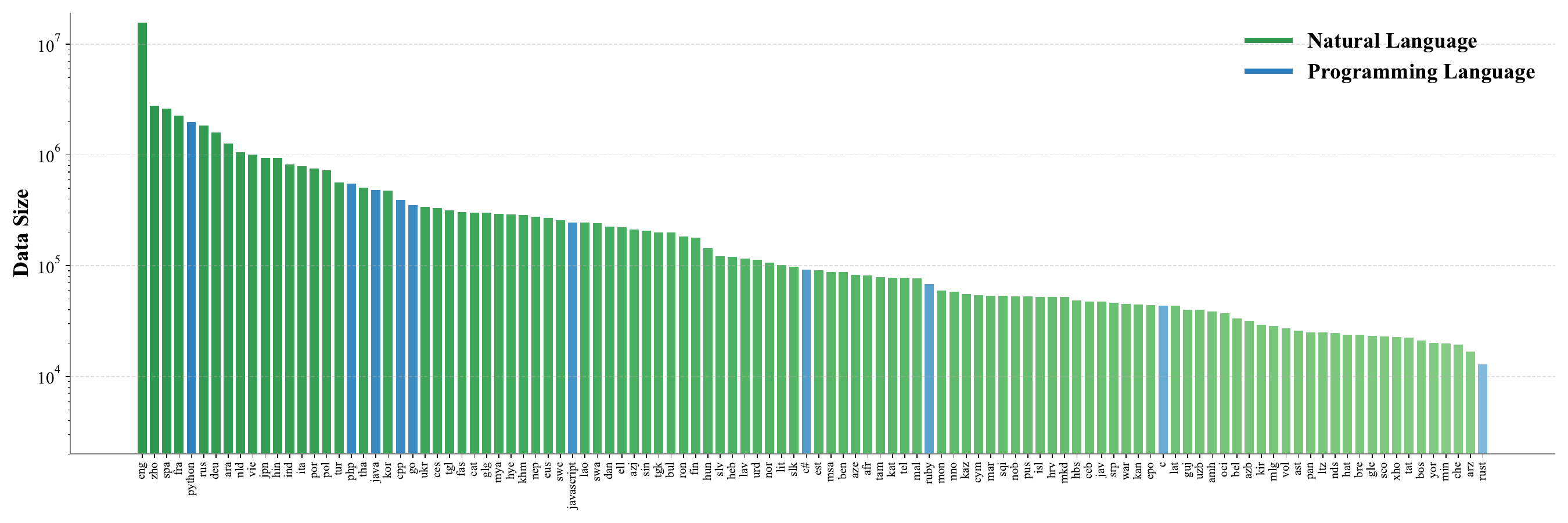}
    \caption{Top-100 natural languages and top-10 programming languages in our training data.}
    \label{fig:language-distribution}
\end{figure*}

A cornerstone of our work is the compilation of a vast and diverse training corpus designed to foster both linguistic inclusivity and broad task competency. We aggregate data from 121 publicly available sources, creating a collection of 50 million training samples that span 282 natural languages (as identified by ISO-639-3 codes) and over 40 programming languages. Crucially, our data curation process is driven by real-world data availability rather than optimizing for specific benchmarks. For instance, our dataset contains substantial data for Spanish and Arabic, which are the 3rd and 7th most represented languages in our corpus (Figure~\ref{fig:language-distribution}), despite these languages lacking dedicated benchmarks in MTEB (see Table~\ref{tab:num-submission}). This approach, which also includes a long tail of low-resource languages and a significant volume of code, aims to build a model with truly global utility and directly contrasts recent open-source datasets such as that released by KaLM-Embedding, which is heavily skewed towards English and Chinese (Figure~\ref{fig:language-comparison}). We provide a more comprehensive linguistic breakdown of our dataset in Appendix~\ref{appendix:data}.

\begin{figure}[t]
    \centering
    \includegraphics[width=1\linewidth]{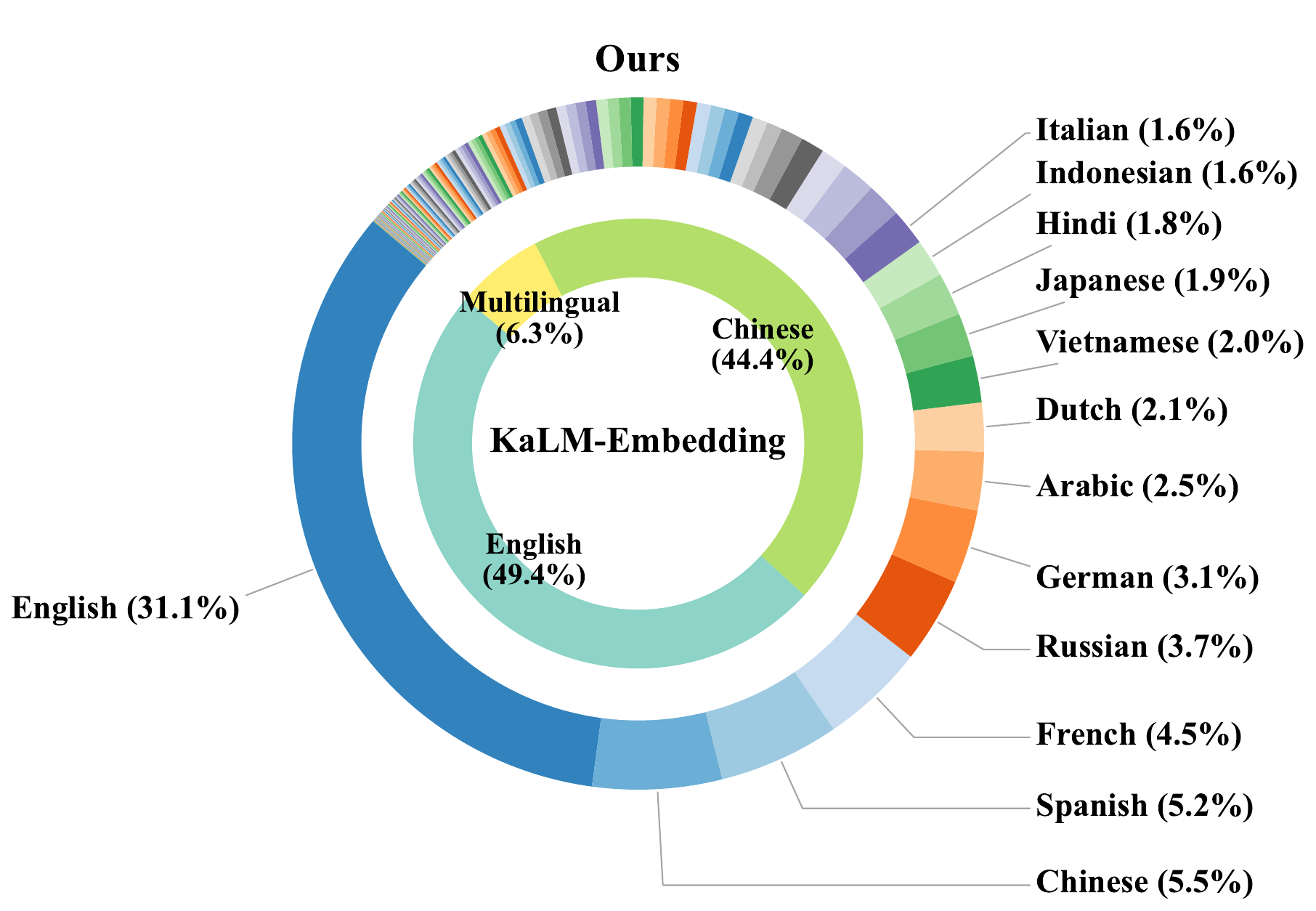}
    \caption{Comparison between the language distribution of our training data (outer circle) and KaLM-Embedding (inner circle). KaLM-Embedding's data is only annotated with three labels, while ours are annotated with specific languages.}
    \label{fig:language-comparison}
\end{figure}

The functional diversity of our dataset is equally critical for training a general-purpose embedding model. As shown in Figure~\ref{fig:task-distribution} in Appendix~\ref{appendix:data}, our collection encompasses a wide spectrum of tasks, ranging from retrieval-focused question answering and bitext mining to classification-oriented sentiment analysis and intent/domain classification.

To leverage this heterogeneity within a unified contrastive learning framework, we follow prior work~\citep{2025NV-Embed,2025f2llm} and consolidate all data into three canonical formats: \emph{retrieval, clustering, and two-way classification}. This consolidation allows the model to learn a versatile embedding space by optimizing a single, consistent objective across disparate data sources and task structures. For the retrieval format, data consists of (query, positive document, hard negatives) tuples. We leverage both in-batch negatives, where other documents in a mini-batch serve as negatives, and explicitly provided hard negatives (mined using Qwen3-Embedding-8B) to create a challenging and efficient training signal. For the clustering format, which also ingests multi-class classification tasks, tuples are formed by sampling an anchor, a positive example from the same class, and a hard negative from a different class. Finally, the two-way classification format directly uses class labels, where a given text serves as the anchor, the corresponding label text is the positive, and the opposite label text is the negative. For both clustering and classification, only hard negatives are utilized to avoid introducing false negatives from in-batch samples.

To maximize the utility of this diverse corpus, we adopt a two-stage training strategy following previous works~\citep{2025NV-Embed,2025Qwen3-Embedding}. The first stage focuses on building a robust semantic foundation by training on a large-scale subset of retrieval datasets, totaling 27 million samples. This phase imbues the model with a strong general understanding of semantic similarity. In the second stage, we conduct fine-tuning on a sampled mixture of 8.3 million samples from all data sources, applying task-specific instructions to the queries. This stage sharpens the model's ability to handle the nuances of diverse downstream applications like classification, reranking, and paraphrase detection.
\vspace{-0.03cm}
\section{Experiments}\label{sec:experiments}
\vspace{-0.05cm}
\subsection{Experimental Setting}

\begin{table*}[th]
    \centering
    \caption{Comparison of our models against previous top-1 and top-5 performance on 17 MTEB benchmark leaderboards. The number of tasks in each benchmark is given in $^{\text{(superscript)}}$. The specific metrics are consistent with the main metrics used by MTEB (e.g., NDCG@10 for retrieval tasks and accuracy for classification tasks).}
    \label{tab:mteb-results}
    \small
    \adjustbox{width=\textwidth,center}{
    \begin{tabular}{rccccccccc}
    \toprule
    \tb{Model} & \tb{Multi.}$^{\text{(131)}}$ & \tb{English}$^{\text{(41)}}$ & \tb{Code}$^{\text{(12)}}$ & \tb{Medical}$^{\text{(12)}}$ & \tb{European}$^{\text{(73)}}$ & \tb{Scan.}$^{\text{(28)}}$ & \tb{Indic}$^{\text{(20)}}$ & \tb{German}$^{\text{(19)}}$ & \tb{French}$^{\text{(25)}}$ \\
    \midrule
    \rowcolor{gray!15} \multicolumn{10}{c}{\textit{Top results on the leaderboard}}\\
    \emph{Top-1} & \tb{72.32} & \tb{75.97} & \tb{80.75} & \tb{66.55} & 63.60 & 65.56 & 70.15 & 59.96 & 70.37 \\
    \emph{Top-5} & 69.45 & 74.61 & 76.00 & 63.83 & 62.32 & 62.01 & 67.39 & 55.72 & 67.25 \\
    \midrule
    \rowcolor{gray!15} \multicolumn{10}{c}{\textit{Ours}}\\
    \emph{8B} & 66.79 & 73.26 & 80.28 & 62.91 & \tb{68.00} & \tb{69.49} & \tb{76.76} & \tb{66.43} & \tb{71.91} \\
    \emph{4B} & 65.80 & 72.89 & 79.95 & 62.05 & 67.53 & 67.88 & 75.15 & 65.49 & 70.97 \\
    \emph{1.7B} & 63.70 & 71.19 & 78.90 & 59.88 & 65.47 & 65.79 & 72.58 & 63.99 & 68.94 \\
    \emph{0.6B} & 61.30 & 70.01 & 77.27 & 57.58 & 63.40 & 62.00 & 66.11 & 61.58 & 66.64 \\
    \emph{330M} & 58.06 & 67.99 & 73.94 & 55.91 & 59.80 & 59.22 & 61.87 & 58.48 & 63.59 \\
    \emph{140M} & 49.81 & 60.35 & 61.78 & 48.28 & 50.40 & 47.95 & 46.37 & 46.66 & 52.83 \\
    \bottomrule
    % \addlinespace[1em] % Creates a visual gap between the two halves
    \rowcolor{gray!30} \multicolumn{10}{c}{\textit{Results continued for remaining languages and average}} \\
    \toprule
    \tb{Model} & \tb{Korean}$^{\text{(6)}}$ & \tb{Polish}$^{\text{(17)}}$ & \tb{Chinese}$^{\text{(32)}}$ & \tb{Japan.}$^{\text{(28)}}$ & \tb{Dutch}$^{\text{(40)}}$ & \tb{Russian}$^{\text{(23)}}$ & \tb{Persian}$^{\text{(52)}}$ & \tb{Viet.}$^{\text{(50)}}$ & \tb{Average} \\
    \midrule
    \rowcolor{gray!15} \multicolumn{10}{c}{\textit{Top results on the leaderboard}}\\
    \emph{Top-1} & \tb{77.01} & 50.95 & \tb{78.52} & 73.18 & 58.38 & \tb{74.16} & \tb{71.58} & 54.74 & 68.46 \\
    \emph{Top-5} &  69.06 & n.a. & 74.87 & 69.51 & 56.23 & 69.39 & 65.26 & 52.37 & 65.95 \\
    \midrule
    \rowcolor{gray!15} \multicolumn{10}{c}{\textit{Ours}}\\
    \emph{8B} & 74.84 & \tb{73.84} & 67.22 & \tb{77.81} & \tb{62.64} & 69.24 & 71.12 & \tb{61.62} & \tb{70.24} \\
    \emph{4B} & 73.32 & 73.14 & 66.55 & 76.65 & 61.42 & 67.93 & 69.94 & 61.20 & 69.29 \\
    \emph{1.7B} & 72.33 & 71.12 & 64.87 & 74.53 & 59.83 & 67.08 & 68.35 & 60.27 & 67.58 \\
    \emph{0.6B} & 68.74 & 68.13 & 62.64 & 71.30 & 56.59 & 63.35 & 65.32 & 57.85 & 64.69 \\
    \emph{330M} & 61.71 & 65.65 & 59.38 & 66.00 & 54.04 & 60.12 & 60.16 & 54.17 & 61.18 \\
    \emph{140M} & 53.07 & 51.22 & 51.02 & 55.97 & 43.89 & 47.24 & 52.50 & 43.67 & 50.76 \\
    \bottomrule
    \end{tabular}
    }
\end{table*}

\vspace{-0.05cm}
\paragraph{Model}
We present a series of 6 models, all trained on identical data in exactly the same order: 140M, 330M, 600M, 1.7B, 4B, 8B. All models are fine-tuned from Qwen3 causal LLMs~\citep{2025Qwen3}, where 600M, 1.7B, 4B, and 8B models correspond to models of the same size in the Qwen3 family, while 140M and 330M models are pruned from the 600M model after training. Following existing embedding models based on the Qwen3 family~\citep{2025Qwen3-Embedding,2025f2llm}, we maintain the causal attention in the models and use EOS token representation as the sequence embedding.

\vspace{-0.11cm}

\paragraph{Training}
Inspired by NV-Embed~\citep{2025NV-Embed} and Qwen3-Embedding~\citep{2025Qwen3-Embedding}, we train the models in two stages. In the first stage, we use only some of the largest retrieval datasets, and do not apply instructions to the data, aiming to inject the causal models with basic capabilities of converting texts into semantic embeddings that are usable by downstream tasks. In the second stage, we sample from all data sources, and apply instructions to queries, enabling a more nuanced understanding of different semantic representation tasks such as retrieval, classification, and paraphrase detection. More details about the amount of training data and hyperparameter settings are given in Appendix~\ref{appendix:training}, where we also demonstrate the empirical benefits of the two-stage training design.

\vspace{-0.11cm}

\paragraph{Evaluation}
We evaluate the models on 17 MTEB benchmarks: Multilingual~\citep{2025MMTEB}, English~\citep{2025MMTEB}, Code~\citep{2025MMTEB}, Medical, European~\citep{2025MMTEB}, Scandinavian~\citep{2024SEB}, Indic~\citep{2025MMTEB}, German~\citep{2023mteb-deu}, French~\citep{2024mteb-fra}, Korean, Polish~\citep{2024mteb-pol}, Chinese~\citep{2023CMTEB}, Japanese~\citep{2026mteb-jpn}, Dutch~\citep{2025mteb-nld}, Russian~\citep{2024mteb-rus}, Persian~\citep{2025mteb-fas}, and Vietnamese~\citep{2025mteb-vie}, totaling 430 tasks across ten types: retrieval, reranking, classification, clustering, pair classification, multilabel classification, STS, instruction reranking, bitext mining, and summarization. More details on these benchmarks and tasks are given in Appendix~\ref{appendix:mteb}. For comparison, we report the previous top-1 score and top-5 score on each benchmark's leaderboard. We also compare with individual models, specifically those from the Qwen3-Embedding~\citep{2025Qwen3-Embedding} and EmbeddingGemma~\citep{2025EmbeddingGemma} families.

\subsection{MTEB Results}

We present the main results in Table~\ref{tab:mteb-results}, comparing the ML-Embed family against the top-performing models on 17 MTEB benchmarks. Our largest model, ML-Embed-8B, establishes new state-of-the-art (SOTA) scores on a remarkable 9 out of 17 benchmarks, demonstrating the effectiveness of our multilingual training corpus.

Critically, these SOTA results are concentrated in benchmarks for languages historically underserved by the research community, directly addressing the linguistic challenge outlined in Section~\ref{sec:introduction}. For instance, on the Polish benchmark, our model achieves a score of 73.84, a staggering +22.89 point improvement over the previous best model. Similarly, we set new records on Vietnamese (+6.88), Indic (+6.61), German (+6.47), Japanese (+4.63), Dutch (+4.26), French (+1.54), and the aggregated benchmarks of Scandinavian (+3.93) and European (+4.40). This demonstrates that our data curation and training methodology successfully produce models with globally equitable performance.

On the highly competitive English and Multilingual benchmarks, our models also perform comparably to the top-5 models on the leaderboard, validating our approach as a strong foundation for general-purpose embeddings. Furthermore, the results exhibit a clear and consistent scaling trend: performance reliably improves with model size across all benchmarks. This indicates that our training recipe is robust and provides a scalable blueprint for developing even more powerful models in the future.

\subsection{Comparison with Similar-Sized Models}\label{sec:model-comparison}

In Table~\ref{tab:mteb-comparison}, we compare our 0.3B and 0.6B models with EmbeddingGemma-0.3B and Qwen3-Embedding-0.6B, respectively. For these two models, we use their public results in the MTEB repository when avaiable, and evaluate the remaining tasks using the same prompts as those used for evaluating our models. The results are similar to those on the MTEB leaderboard: our models demonstrate superior performance on the less-attended Code, Scandinavian, German, French, Korean, Polish, Japanese, Dutch, and Vietnamese benchmarks, while underperforming on English, Chinese, and Multilingual benchmarks.

\begin{table*}[th]
    \centering
    \caption{Comparison of our models with EmbeddingGemma and Qwen3-Embedding. The number of tasks in each benchmark is given in $^{\text{(superscript)}}$.}
    \label{tab:mteb-comparison}
    \small
    \adjustbox{width=\textwidth,center}{
    \begin{tabular}{lccccccccc}
    \toprule
    Model & Multi.$^{\text{(131)}}$ & English$^{\text{(41)}}$ & Code$^{\text{(12)}}$ & Medical$^{\text{(12)}}$ & European$^{\text{(73)}}$ & Scan.$^{\text{(28)}}$ & Indic$^{\text{(20)}}$ & German$^{\text{(19)}}$ & French$^{\text{(25)}}$ \\
    \midrule
    \rowcolor{gray!15} \multicolumn{10}{c}{\textit{0.3B}}\\
    \emph{EmbedGemma} & \ud{61.15} & \ud{69.67} & 68.76 & 51.24 & \ud{62.50} & 54.39 & \ud{66.11} & 56.28 & 61.90 \\
    \emph{Ours} & 58.06 & 67.99 & \ud{73.94} & \ud{55.91} & 59.80 & \ud{59.22} & 61.87 & \ud{58.48} & \ud{63.59} \\
    \midrule
    \rowcolor{gray!15} \multicolumn{10}{c}{\textit{0.6B}}\\
    \emph{Qwen3-Embed} & \ud{64.34} & \ud{70.47} & 75.42 & \ud{60.16} & \ud{63.91} & 60.99 & \ud{66.53} & 59.45 & 63.01 \\
    \emph{Ours} & 61.30 & 70.01 & \ud{77.27} & 57.58 & 63.40 & \ud{62.00} & 66.11 & \ud{61.58} & \ud{66.64} \\
    % \midrule
    % \rowcolor{gray!15} \multicolumn{10}{c}{\textit{4B}}\\
    % \emph{Qwen3-Embed} & 69.45 & 74.61 & 80.07 & 65.08 & 69.44 & 68.56 & 75.27 & 65.25 & 70.54 \\
    % \emph{Ours} &  \\
    % \midrule
    % \rowcolor{gray!15} \multicolumn{10}{c}{\textit{8B}}\\
    % \emph{Qwen3-Embed} & 70.58 & 75.23 & 80.69 & 68.73 & 70.24 & 70.15 & 77.09 & 66.57 & 71.84 \\
    % \emph{Ours} &  \\
    \bottomrule
    % \addlinespace[1em] % Creates a visual gap between the two halves
    \rowcolor{gray!30} \multicolumn{10}{c}{\textit{Results continued for remaining languages and average}} \\
    \toprule
    Model & Korean$^{\text{(6)}}$ & Polish$^{\text{(17)}}$ & Chinese$^{\text{(32)}}$ & Japan.$^{\text{(28)}}$ & Dutch$^{\text{(40)}}$ & Russian$^{\text{(23)}}$ & Persian$^{\text{(52)}}$ & Viet.$^{\text{(50)}}$ & \tb{Avg.} \\
    \midrule
    \rowcolor{gray!15} \multicolumn{10}{c}{\textit{0.3B}}\\
    \emph{EmbedGemma} & 58.24 & 64.70 & 50.40 & 60.82 & 50.98 & \ud{64.57} & \ud{67.11} & 43.45 & 59.55 \\
    \emph{Ours} & \ud{61.71} & \ud{65.65} & \ud{59.38} & \ud{66.00} & \ud{54.04} & 60.12 & 60.16 & \ud{54.17} & \ud{61.18} \\
    \midrule
    \rowcolor{gray!15} \multicolumn{10}{c}{\textit{0.6B}}\\
    \emph{Qwen3-Embed} & 65.29 & 67.42 & \ud{66.71} & 67.28 & 54.27 & \ud{64.20} & 62.88 & 56.01 & 64.02 \\
    \emph{Ours} &  \ud{68.74} & \ud{68.13} & 62.64 & \ud{71.30} & \ud{56.59} & 63.35 & \ud{65.32} & \ud{57.85} & \ud{64.69} \\
    % \midrule
    % \rowcolor{gray!15} \multicolumn{10}{c}{\textit{4B}}\\
    % \emph{Qwen3-Embed} &  67.57 & 74.69 & 72.53 & 71.17 & 62.00 & 69.89 & 69.25 & 60.15 & 69.74 \\
    % \emph{Ours} &  \\
    % \midrule
    % \rowcolor{gray!15} \multicolumn{10}{c}{\textit{8B}}\\
    % \emph{Qwen3-Embed} & 68.17 & 75.27 & 74.08 & 73.74 & 63.37 & 71.07 & 69.84 & 61.61 & 71.07 \\
    % \emph{Ours} &  \\
    \bottomrule
    \end{tabular}
    }
\end{table*}

\subsection{Ablation Studies}

To dissect the individual contributions of our proposed efficiency methods, we conduct a series of ablation studies on the English subset using the 0.6B model.

\begin{figure}[t]
    \centering
    \includegraphics[width=0.85\linewidth]{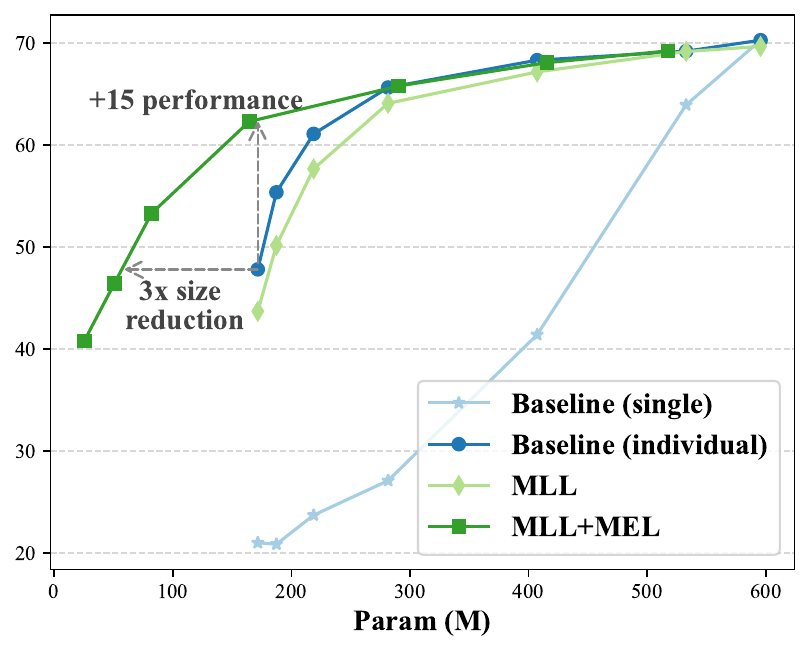}
    \caption{Ablation results of models pruned to different depths on MTEB-English. Each point on the baseline (individual) curve represents an individual trained model, while points on the Baseline (single), MLL, and MLL+MEL curves are models of different depths pruned from a single trained model.}
    \label{fig:results_mll}
\end{figure}

\paragraph{MLL and MEL Synergy}
First, we investigate the interplay between Matryoshka Layer Learning (MLL) and Matryoshka Embedding Learning (MEL). As shown in Figure~\ref{fig:results_mll}, we compare four settings: (1) a single baseline, evaluated at different exit layers; (2) baselines of individually trained models at different depths; (3) a single model trained with MLL and evaluated at different exit layers; and (4) a fourth model combining MLL and MEL. MLL alone presents a classic trade-off: it enables training a single, depth-flexible model for the computational cost of one, but the resulting shallower models slightly underperform individually trained counterparts. However, the introduction of MEL dramatically alters this dynamic. By significantly reducing the parameter count of the embedding layer, MEL allows for a much deeper model at the same parameter budget. For example, our MLL+MEL model with 4 layers has the same parameter count (~170M) as a 1-layer baseline model but achieves a 15-point higher score. At equivalent performance levels, the MLL+MEL model is 3x smaller, confirming the powerful synergy between these two techniques for creating parameter-efficient models.

\begin{figure}[t]
    \centering
    \includegraphics[width=0.82\linewidth]{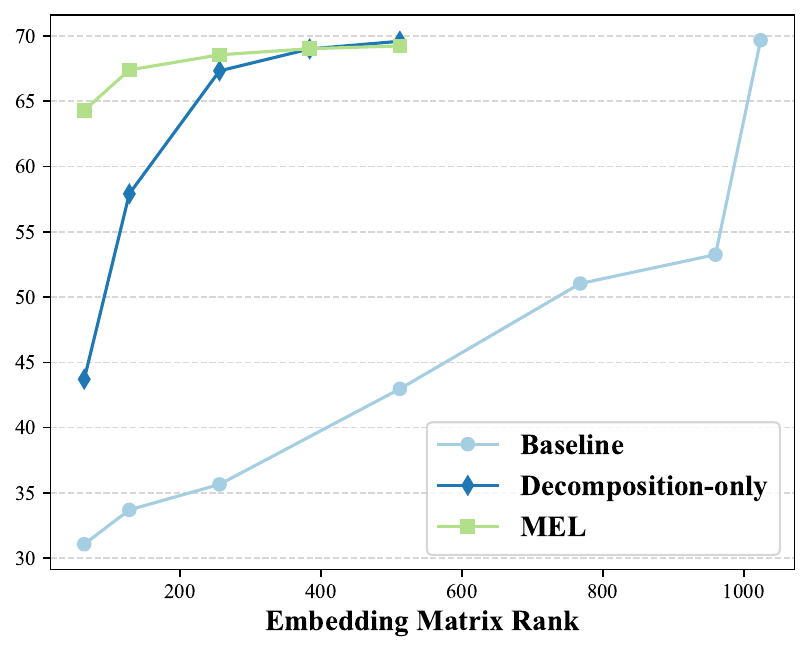}
    \caption{Ablation results of decomposing the embedding layer to varying ranks at inference time. Baseline model is trained with the original embedding. Decomposition-only model is trained with decomposed embedding (at rank 512). MEL model is trained with decomposed embedding plus Matryoshka Embedding Learning.}
    \label{fig:results_mel}
\end{figure}

\paragraph{Robustness of MEL}
Next, we isolate the effect of MEL on inference-time compression. We compare a baseline model against two variants: one trained with a factorized embedding layer at rank 512 (``Decomposition-only'') and another additionally trained with the nested rank objective of MEL. At inference, we apply SVD to each model's embedding matrix and evaluate performance at progressively smaller ranks. The results in Figure~\ref{fig:results_mel} are stark. The baseline model is extremely brittle - its performance collapses catastrophically (from 69.68 to 53.25) with even minor rank reduction (from 1024 to 960). The decomposition-only model is more robust, as the low-rank structure acts as a training regularizer even though the model is not trained to concentrate information on the leading ranks. Notably, it achieves almost identical performance to the baseline (69.60) at rank 512, demonstrating the redundancy in the embedding matrix. The MEL-trained model demonstrates superior robustness against decomposition, declining much more slowly as the rank diminishes, retaining a strong score of 64.30 even when compressed to a rank of just 64. This confirms that MEL is highly effective at producing models that are robust to aggressive, post-hoc compression.

\begin{table*}[th]
    \centering
    \caption{Comparison of baseline training, pruned model training, and 3D-ML on EuroBERT backbone.}
    \label{tab:eurobert}
    \small
    \adjustbox{width=\textwidth,center}{
    \begin{tabular}{lccccccccc}
    \toprule
    Model & Multi. & English & Code & Medical & European & Scan. & Indic & German & French \\
    \midrule
    \emph{Baseline (210M)} & 59.86 & 68.52 & 64.03 & 55.93 & 59.83 & 54.43 & 55.36 & 62.90 & 62.90 \\
    \emph{Pruned (120M)} & 46.08 & 54.65 & 35.44 & 39.52 & 44.79 & 40.75 & 42.90 & 47.69 & 47.94 \\
    \emph{3D-ML (120M)} & 56.94 & 66.13 & 59.76 & 51.83 & 56.75 & 51.36 & 50.53 & 59.84 & 58.96 \\
    \bottomrule
    \rowcolor{gray!30} \multicolumn{10}{c}{\textit{Results continued for remaining languages and average}} \\
    \toprule
    Model & Korean & Polish & Chinese & Japan. & Dutch & Russian & Persian & Viet. & \tb{Avg.} \\
    \midrule
    \emph{Baseline (210M)} & 61.01 & 64.63 & 59.37 & 71.82 & 52.26 & 58.44 & 60.00 & 55.22 & 60.38 \\
    \emph{Pruned (120M)} & 42.44 & 47.73 & 43.48 & 52.86 & 35.39 & 42.29 & 48.30 & 37.48 & 44.10 \\
    \emph{3D-ML (120M)} & 58.27 & 59.13 & 56.18 & 67.56 & 47.84 & 54.50 & 58.24 & 51.24 & 56.77 \\
    \bottomrule
    \end{tabular}
    }
\end{table*}

\begin{figure}[th]
    \centering
    \includegraphics[width=0.8\linewidth]{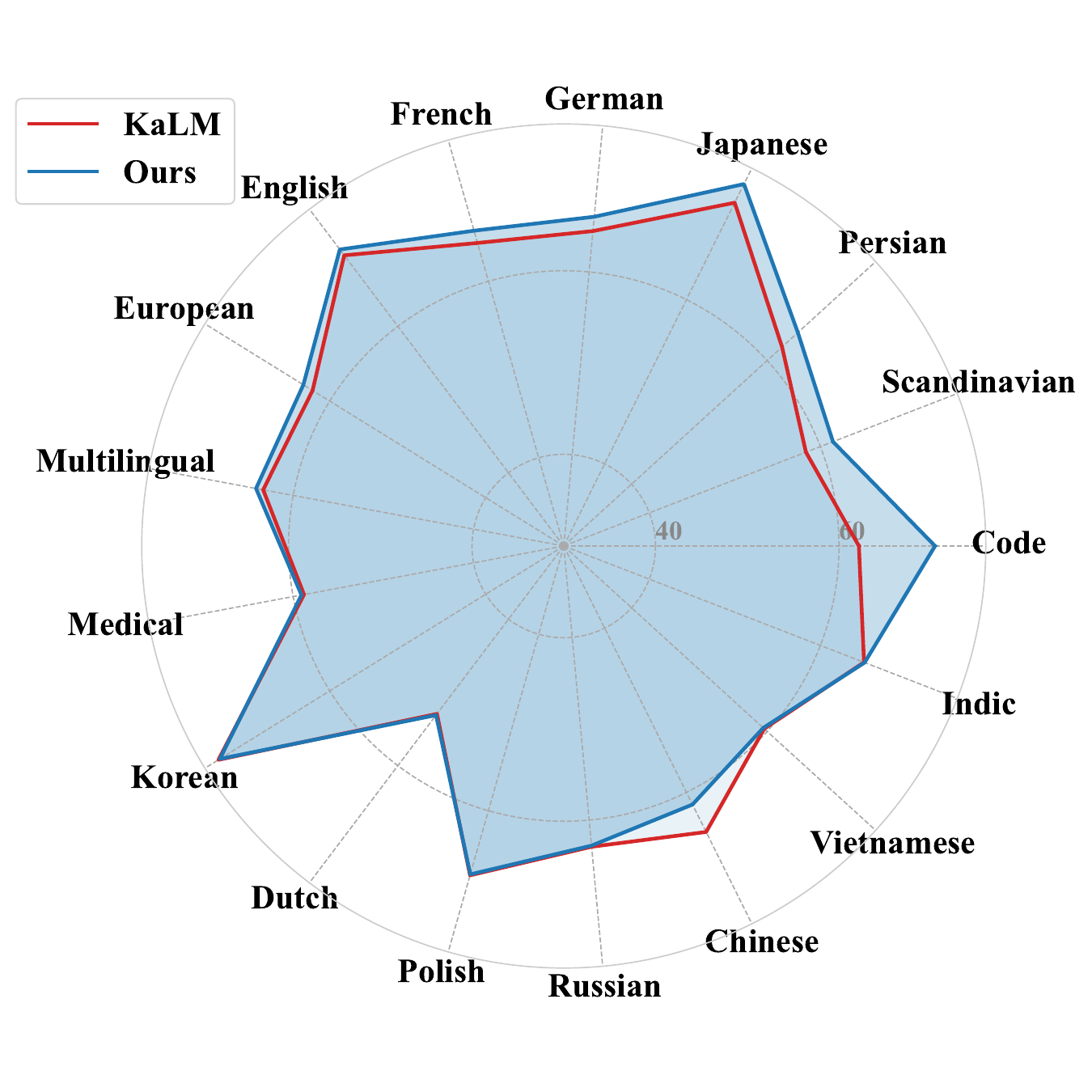}
    \caption{Comparison between 0.6B models trained on our data and KaLM-Embedding data.}
    \label{fig:data_ablation}
\end{figure}

\vspace{-0.1cm}
\paragraph{Data Comparison}
To isolate the impact of our curated training corpus, we conduct a head-to-head comparison against the recently released KaLM-Embedding finetuning data~\citep{2025KaLM-Embedding-V2}. Starting from the same stage-1 checkpoint, we finetune two identical 0.6B models using our stage-2 data and the similar-sized KaLM-Embedding data, respectively. The results, shown in Figure~\ref{fig:data_ablation}, demonstrate the distinct advantages of our data curation strategy. Our model achieves superior performance on 9 out of 17 benchmarks, including the composite Multilingual, European, and Scandinavian sets, as well as English, French, German, Japanese, and Persian. The most significant lead is on the Code benchmark, highlighting our data's wider domain coverage. While the KaLM-Embedding data produces a stronger model for Chinese - an expected outcome given its heavy concentration on Chinese and English data (Figure~\ref{fig:language-comparison}) - our dataset achieves on-par performance across seven other benchmarks, including Korean, Polish, Dutch, Indic, Russian, and Vietnamese. This outcome confirms that our focus on linguistic diversity yields a more globally robust model, trading hyper-specialization in a single language for broader competence.

\paragraph{Generalization to Different Backbones}
To verify the effectiveness of 3D-ML, we conduct additional experiments using EuroBERT-210M~\citep{2025eurobert} as the backbone. We train three models on the stage-2 data described in Section~\ref{sec:data}: 1) baseline finetuning, 2) structural pruning to 120M parameters followed by finetuning, and 3) 3D-ML training followed by structural pruning to 120M parameters. The results in Table~\ref{tab:eurobert} show that training with 3D-ML leads to a minimal performance drop compared with naive structural pruning, demonstrating the generalizability of 3D-ML.

In Appendix~\ref{appendix:additional-results}, we provide additional experiments including 1) applying 3D-ML to training on specific languages, 2) applying MEL to language modeling, and 3) in-depth analysis of efficiency gains.
\section{Conclusion}

This paper addresses the critical challenges of computational cost, linguistic bias, and lack of transparency hindering the development of text embeddings. We introduce ML-Embed, a family of models built using our 3-Dimensional Matryoshka Learning (3D-ML) framework. 3D-ML integrates Matryoshka Embedding (MEL), Layer (MLL), and Representation (MRL) Learning to achieve comprehensive efficiency across the model lifecycle. Paired with a newly curated, massively multilingual open-source dataset, our approach demonstrates that efficiency and inclusivity can yield state-of-the-art performance. Our 8B model sets new records on 9 of 17 MTEB benchmarks, with dramatic improvements in historically understudied languages such as Polish (+22.89) and Vietnamese (+6.88). Ablation studies confirm the synergistic benefits of 3D-ML components, creating powerful models adaptable to diverse computational budgets. The studies also highlight the superiority of our curated data.

By releasing our models, data, and code, we offer a reproducible blueprint for building globally equitable and efficient AI systems, dismantling the transparency barrier. Our work paves the way for future research in scaling these techniques to larger models, expanding linguistic coverage, and deploying powerful embeddings on resource-constrained devices. We hope this work steers the field toward a more inclusive and accessible future for text representation learning.

% Beyond the benchmark scores, our work provides a practical and reproducible blueprint for building globally equitable and computationally efficient AI systems. By releasing all models, data, and code, we dismantle the transparency barrier and invite the community to build upon, audit, and extend our methods. Future work can explore scaling these techniques to even larger models, further expanding the linguistic and domain coverage of the training data, and investigating novel applications for efficient embeddings on edge devices. Ultimately, we hope to steer the field towards a more inclusive, transparent, and accessible future for text representation learning.

% \section*{Acknowledgements}
% This work was supported by Ant Group.

% \textbf{Do not} include acknowledgements in the initial version of the paper

% If a paper is accepted, the final camera-ready version can (and usually should)
% include acknowledgements.  Such acknowledgements should be placed at the end of
% the section, in an unnumbered section that does not count towards the paper
% page limit. Typically, this will include thanks to reviewers who gave useful
% comments, to colleagues who contributed to the ideas, and to funding agencies
% and corporate sponsors that provided financial support.

\section*{Impact Statement}

This paper presents work whose goal is to advance the field of Machine
Learning. There are many potential societal consequences of our work, none
which we feel must be specifically highlighted here.


\bibliography{example_paper}
\bibliographystyle{icml2026}

%%%%%%%%%%%%%%%%%%%%%%%%%%%%%%%%%%%%%%%%%%%%%%%%%%%%%%%%%%%%%%%%%%%%%%%%%%%%%%%
%%%%%%%%%%%%%%%%%%%%%%%%%%%%%%%%%%%%%%%%%%%%%%%%%%%%%%%%%%%%%%%%%%%%%%%%%%%%%%%
% APPENDIX
%%%%%%%%%%%%%%%%%%%%%%%%%%%%%%%%%%%%%%%%%%%%%%%%%%%%%%%%%%%%%%%%%%%%%%%%%%%%%%%
%%%%%%%%%%%%%%%%%%%%%%%%%%%%%%%%%%%%%%%%%%%%%%%%%%%%%%%%%%%%%%%%%%%%%%%%%%%%%%%
\newpage
\appendix
\onecolumn
\section{Details on Training Data}\label{appendix:data}

\begin{table}[th]
    \centering
    \caption{Natural language distribution in our training data (part1).}
    \label{tab:language-distribution-1}
    \small
    \adjustbox{width=\textwidth-1cm,center}{
    \begin{tabular}{m{1cm}<{\centering}cr|m{1cm}<{\centering}cr|m{1cm}<{\centering}cr}
    \toprule
\tb{ISO Code} & \tb{Language} & \tb{Samples} & \tb{ISO Code} & \tb{Language} & \tb{Samples} & \tb{ISO Code} & \tb{Language} & \tb{Samples} \\
    \cmidrule(r){1-3}\cmidrule(lr){4-6}\cmidrule(l){7-9}
eng & English & 15,683,866 & slk & Slovak & 97,974 & tat & Tatar & 22,327 \\
zho & Chinese & 2,791,623 & est & Estonian & 91,102 & bos & Bosnian & 21,175 \\
spa & Spanish & 2,607,192 & msa & Malay & 87,873 & yor & Yoruba & 20,139 \\
fra & French & 2,251,567 & ben & Bengali & 87,787 & min & Minangkabau & 19,868 \\
rus & Russian & 1,848,954 & aze & Azerbaijani & 82,209 & che & Chechen & 19,518 \\
deu & German & 1,593,329 & afr & Afrikaans & 81,233 & arz & Egyptian Arabic & 16,783 \\
ara & Arabic & 1,264,371 & tam & Tamil & 78,384 & lmo & Lombard & 16,575 \\
nld & Dutch & 1,051,914 & kat & Georgian & 77,567 & arg & Aragonese & 16,500 \\
vie & Vietnamese & 1,005,512 & tel & Telugu & 77,362 & bak & Bashkir & 16,451 \\
jpn & Japanese & 935,603 & mal & Malayalam & 76,518 & som & Somali & 16,369 \\
hin & Hindi & 931,761 & mon & Mongolian & 59,851 & als & Tosk Albanian & 15,655 \\
ind & Indonesian & 817,131 & nno & Norwegian Nynorsk & 58,298 & ido & Ido & 15,613 \\
ita & Italian & 790,507 & kaz & Kazakh & 55,317 & szl & Silesian & 14,845 \\
por & Portuguese & 752,642 & cym & Welsh & 53,951 & wuu & Wu Chinese & 14,762 \\
pol & Polish & 730,068 & mar & Marathi & 53,803 & new & Nepal Bhasa & 14,714 \\
tur & Turkish & 565,158 & sqi & Albanian & 53,602 & chv & Chuvash & 13,759 \\
tha & Thai & 503,612 & nob & Norwegian Bokmål & 52,905 & pnb & Western Panjabi & 13,657 \\
kor & Korean & 478,665 & pus & Pushto & 52,753 & fry & Western Frisian & 13,649 \\
ukr & Ukrainian & 337,739 & isl & Icelandic & 52,546 & snd & Sindhi & 13,210 \\
ces & Czech & 330,256 & hrv & Croatian & 52,505 & ori & Oriya & 12,791 \\
tgl & Tagalog & 313,985 & mkd & Macedonian & 52,504 & plt & Plateau Malagasy & 12,717 \\
fas & Persian & 302,869 & hbs & Serbo-Croatian & 48,523 & scn & Sicilian & 12,694 \\
cat & Catalan & 301,984 & ceb & Cebuano & 47,408 & kur & Kurdish & 11,872 \\
glg & Galician & 301,386 & jav & Javanese & 47,283 & sun & Sundanese & 11,793 \\
mya & Burmese & 294,167 & srp & Serbian & 46,108 & bar & Bavarian & 11,193 \\
hye & Armenian & 288,622 & war & Waray & 45,348 & yid & Yiddish & 10,785 \\
khm & Khmer & 287,530 & kan & Kannada & 44,534 & ckb & Central Kurdish & 9,829 \\
nep & Nepali & 276,057 & epo & Esperanto & 44,266 & fao & Faroese & 9,825 \\
eus & Basque & 270,551 & lat & Latin & 43,335 & ina & Interlingua & 9,782 \\
swe & Swedish & 256,469 & guj & Gujarati & 40,184 & gla & Scottish Gaelic & 9,769 \\
lao & Lao & 244,750 & uzb & Uzbek & 39,951 & bug & Buginese & 9,662 \\
swa & Swahili & 241,497 & amh & Amharic & 38,763 & que & Quechua & 9,406 \\
dan & Danish & 224,504 & oci & Occitan & 37,413 & bpy & Bishnupriya & 9,400 \\
ell & Modern Greek & 223,254 & bel & Belarusian & 33,330 & san & Sanskrit & 8,730 \\
azj & North Azerbaijani & 213,046 & azb & South Azerbaijani & 31,815 & lim & Limburgan & 8,573 \\
sin & Sinhala & 207,903 & kir & Kirghiz & 29,319 & hau & Hausa & 8,435 \\
tgk & Tajik & 200,631 & mlg & Malagasy & 28,661 & mai & Maithili & 8,180 \\
bul & Bulgarian & 200,127 & vol & Volapük & 27,187 & zsm & Standard Malay & 8,179 \\
ron & Romanian & 184,034 & ast & Asturian & 26,004 & ibo & Igbo & 8,132 \\
fin & Finnish & 178,810 & pan & Panjabi & 25,096 & vec & Venetian & 8,121 \\
hun & Hungarian & 144,561 & ltz & Luxembourgish & 25,092 & ilo & Iloko & 7,968 \\
slv & Slovenian & 122,065 & nds & Low German & 24,713 & asm & Assamese & 7,042 \\
heb & Hebrew & 119,685 & hat & Haitian & 23,940 & sah & Yakut & 7,011 \\
lav & Latvian & 115,288 & bre & Breton & 23,931 & arb & Standard Arabic & 6,945 \\
urd & Urdu & 113,258 & gle & Irish & 23,148 & sna & Shona & 6,933 \\
nor & Norwegian & 107,023 & sco & Scots & 23,032 & mlt & Maltese & 6,911 \\
lit & Lithuanian & 101,937 & xho & Xhosa & 22,799 & zul & Zulu & 6,669 \\
    \bottomrule
    \end{tabular}
    }
\end{table}

\begin{table}[th]
    \centering
    \caption{Natural language distribution in our training data (part2).}
    \label{tab:language-distribution-2}
    \small
    \adjustbox{width=\textwidth-1cm,center}{
    \begin{tabular}{m{1cm}<{\centering}cr|m{1cm}<{\centering}cr|m{1cm}<{\centering}cr}
    \toprule
ISO Code & Language & Samples & ISO Code & Language & Samples & ISO Code & Language & Samples \\
    \cmidrule(r){1-3}\cmidrule(lr){4-6}\cmidrule(l){7-9}
mzn & Mazanderani & 6,352 & tsn & Tswana & 1,539 & lvs & Standard Latvian & 800 \\
uig & Uighur & 6,190 & mwl & Mirandese & 1,491 & mag & Magahi & 800 \\
oss & Iron Ossetic & 5,893 & div & Dhivehi & 1,387 & mni & Manipuri & 800 \\
tuk & Turkmen & 5,854 & kbp & Kabiyè & 1,349 & mos & Mossi & 800 \\
ary & Moroccan Arabic & 5,703 & chm & Mari & 1,238 & nqo & N'Ko & 800 \\
wln & Walloon & 5,408 & ewe & Ewe & 1,220 & nus & Nuer & 800 \\
cdo & Min Dong Chinese & 5,175 & smo & Samoan & 1,175 & ory & Odia & 800 \\
npi & Nepali & 5,156 & tso & Tsonga & 1,174 & prs & Dari & 800 \\
nap & Neapolitan & 4,778 & fij & Fijian & 1,122 & quy & Ayacucho Quechua & 800 \\
ace & Achinese & 4,758 & bam & Bambara & 1,061 & sat & Santali & 800 \\
mrj & Western Mari & 4,728 & lin & Lingala & 1,046 & tpi & Tok Pisin & 800 \\
xmf & Mingrelian & 4,712 & nav & Navajo & 1,028 & tum & Tumbuka & 800 \\
pes & Iranian Persian & 4,414 & roh & Romansh & 999 & tzm & Central Atlas Tamazight & 800 \\
diq & Dimli & 4,140 & ssw & Swati & 982 & umb & Umbundu & 800 \\
apc & Levantine Arabic & 4,079 & awa & Awadhi & 954 & uzn & Northern Uzbek & 800 \\
wol & Wolof & 4,068 & pag & Pangasinan & 954 & ydd & Eastern Yiddish & 800 \\
pbt & Southern Pashto & 3,796 & cor & Cornish & 938 & yue & Yue Chinese & 800 \\
nso & Pedi & 3,676 & dzo & Dzongkha & 928 & dyu & Dyula & 799 \\
srd & Sardinian & 3,614 & udm & Udmurt & 890 & lua & Luba-Lulua & 799 \\
ban & Balinese & 3,581 & fon & Fon & 883 & twi & Twi & 798 \\
lij & Ligurian & 3,487 & kon & Kongo & 873 & aeb & Tunisian Arabic & 793 \\
hsb & Upper Sorbian & 3,440 & glv & Manx & 864 & kik & Kikuyu & 761 \\
acq & Ta'izzi-Adeni Arabic & 3,188 & tir & Tigrinya & 851 & tyv & Tuvinian & 626 \\
crh & Crimean Tatar & 2,985 & pms & Piemontese & 842 & ava & Avaric & 588 \\
mri & Maori & 2,922 & myv & Erzya & 840 & aym & Aymara & 587 \\
egl & Emilian & 2,859 & sag & Sango & 827 & krc & Karachay-Balkar & 587 \\
ars & Najdi Arabic & 2,712 & run & Rundi & 825 & ful & Fulah & 581 \\
grn & Guarani & 2,705 & acm & Mesopotamian Arabic & 800 & orm & Oromo & 548 \\
nya & Chichewa & 2,607 & aka & Akan & 800 & stq & Saterfriesisch & 461 \\
hif & Fiji Hindi & 2,566 & ayr & Central Aymara & 800 & lah & Lahnda & 450 \\
kas & Kashmiri & 2,363 & bem & Bemba & 800 & ton & Tonga & 391 \\
fur & Friulian & 2,284 & bho & Bhojpuri & 800 & mdf & Moksha & 314 \\
swh & Swahili & 2,253 & cjk & Chokwe & 800 & haw & Hawaiian & 299 \\
fil & Filipino & 2,156 & dik & Southwestern Dinka & 800 & nia & Nias & 297 \\
sme & Northern Sami & 2,156 & fuv & Nigerian Fulfulde & 800 & bis & Bislama & 272 \\
shn & Shan & 2,098 & gaz & West Central Oromo & 800 & alt & Southern Altai & 250 \\
sot & Southern Sotho & 2,089 & hne & Chhattisgarhi & 800 & srn & Sranan Tongo & 204 \\
kin & Kinyarwanda & 2,031 & kab & Kabyle & 800 & ven & Venda & 194 \\
lug & Ganda & 1,994 & kac & Kachin & 800 & kbd & Kabardian & 172 \\
pap & Papiamento & 1,974 & kam & Kamba & 800 & xal & Kalmyk & 122 \\
cos & Corsican & 1,949 & kea & Kabuverdianu & 800 & din & Dinka & 104 \\
mhr & Eastern Mari & 1,633 & khk & Halh Mongolian & 800 & jam & Jamaican Creole English & 100 \\
bjn & Banjar & 1,600 & kmb & Kimbundu & 800 & kal & Kalaallisut & 92 \\
knc & Central Kanuri & 1,600 & kmr & Northern Kurdish & 800 & iku & Inuktitut & 84 \\
taq & Tamasheq & 1,600 & ltg & Latgalian & 800 & guc & Wayuu & 52 \\
kom & Komi & 1,583 & luo & Luo & 800 & chr & Cherokee & 51 \\
bod & Tibetan & 1,563 & lus & Lushai & 800 & ady & Adyghe & 33 \\
    \bottomrule
    \end{tabular}
    }
\end{table}

\begin{table}[th]
    \centering
    \caption{Programming language distribution in our training data.}
    \label{tab:language-distribution-3}
    \small
    % \adjustbox{width=\textwidth-1cm,center}{
    \begin{tabular}{cr|cr}
    \toprule
Language & Samples & Language & Samples \\
    \cmidrule(r){1-2}\cmidrule(lr){3-4}
python & 1,972,390 & css & 1,003 \\
php & 553,651 & typescript & 888 \\
java & 483,469 & r & 636 \\
cpp & 393,514 & lisp & 467 \\
go & 351,586 & jsx & 436 \\
javascript & 245,632 & objective-c & 327 \\
c\# & 92,008 & json & 264 \\
ruby & 68,317 & xml & 258 \\
c & 43,487 & yaml & 180 \\
rust & 12,924 & assembly & 171 \\
kotlin & 11,284 & powershell & 162 \\
sql & 6,826 & vba & 157 \\
pascal & 6,299 & lua & 126 \\
d & 5,278 & matlab & 114 \\
haskell & 4,967 & dart & 107 \\
scala & 4,120 & bash & 105 \\
html & 2,777 & http & 99 \\
shell & 2,095 & graphql & 89 \\
perl & 2,009 & svg & 82 \\
swift & 1,907 & vb.net & 75 \\
ocaml & 1,894 & groovy & 63 \\
csharp & 1,891 & Misc. & 2,301 \\
    \bottomrule
    \end{tabular}
    % }
\end{table}

\begin{table}[th]
    \caption{Number of samples in our collected training dataset (part 1).}
    \label{tab:data-1}
    \tiny
    \centering
    \adjustbox{width=\textwidth,center}{
    \begin{tabular}{lccrm{9cm}}
    \toprule
        \tb{Name} & \tb{Language} & \tb{Format} & \tb{Size} & \tb{URL} \\
    \midrule
\rowcolor{gray!15} \multicolumn{5}{c}{Bitext Mining}\\
UNPC~\citep{2016unpc} & 6 & Retrieval & 2,922,245 & \url{huggingface.co/datasets/Helsinki-NLP/un_pc} \\
ParaCrawl~\citep{2020paracrawl} & 30 & Retrieval & 10,684,184 & \url{paracrawl.eu/index.php} \\
BactrianX Translation~\citep{2023bactrianx} & 52 & Clustering & 491,282 & \url{huggingface.co/datasets/MBZUAI/Bactrian-X} \\
Europarl~\citep{2005europarl} & 21 & Clustering & 477,566 & \url{huggingface.co/datasets/Helsinki-NLP/europarl} \\
\midrule
\rowcolor{gray!15} \multicolumn{5}{c}{Question Answering}\\
WebFAQ~\citep{2025webfaq} & 49 & Retrieval & 4,368,504 & \url{huggingface.co/datasets/PaDaS-Lab/webfaq-retrieval} \\
mMARCO~\citep{2021mmarco} & 14 & Retrieval & 5,470,174 & \url{huggingface.co/datasets/unicamp-dl/mmarco} \\
PAQ~\citep{2021PAQ} & en & Retrieval & 938,771 & \url{huggingface.co/datasets/sentence-transformers/paq} \\
SQuAD~\citep{2016SQuAD} & en & Retrieval & 89,509 & \url{huggingface.co/datasets/rajpurkar/squad} \\
Stack Exchange~\citep{2021StackExchangeDataset} & en & Retrieval & 754,705 & \url{huggingface.co/datasets/flax-sentence-embeddings/stackexchange_titlebody_best_voted_answer_jsonl} \\
Arguana~\citep{2018Arguana} & en & Retrieval & 22,848 & \url{huggingface.co/datasets/BeIR/arguana-generated-queries} \\
Natural Questions~\citep{2019NaturalQuestions} & en & Retrieval & 97,209 & \url{huggingface.co/datasets/sentence-transformers/natural-questions} \\
HotpotQA~\citep{2018HotpotQA} & en & Retrieval & 120,528 & \url{huggingface.co/datasets/mteb/hotpotqa} \\
ELI5~\citep{2019ELI5} & en & Retrieval & 161,345 & \url{huggingface.co/datasets/Pavithree/eli5} \\
FiQA2018~\citep{2018FIQA} & en & Retrieval & 7,452 & \url{huggingface.co/datasets/mteb/fiqa} \\
BioASQ~\citep{2015BioASQ} & en & Retrieval & 125,248 & \url{huggingface.co/datasets/BeIR/bioasq-generated-queries} \\
NFCorpus~\citep{2016NFCorpus} & en & Retrieval & 1,283 & \url{huggingface.co/datasets/mteb/nfcorpus} \\
TriviaQA~\citep{2017TriviaQA} & en & Retrieval & 60,025 & \url{huggingface.co/datasets/sentence-transformers/trivia-qa-triplet} \\
PubMedQA~\citep{2019PubMedQA} & en & Retrieval & 60,227 & \url{huggingface.co/datasets/qiaojin/PubMedQA} \\
Amazon QA~\citep{2019AmazonQA} & en & Retrieval & 59,340 & \url{github.com/amazonqa/amazonqa} \\
MIRACL~\citep{2023MIRACL} & 16 & Retrieval & 26,740 & \url{huggingface.co/datasets/miracl/miracl} \\
Mr.TyDi~\citep{2021mrtidy} & 11 & Retrieval & 48,619 & \url{huggingface.co/datasets/mteb/mrtidy} \\
MLDR~\citep{2024BGE-M3} & 13 & Retrieval & 40,264 & \url{huggingface.co/datasets/Shitao/MLDR} \\
MKQA~\citep{2021mkqa} & 26 & Retrieval & 69,287 & \url{huggingface.co/datasets/mteb/MKQARetrieval} \\
StackOverflowQA~\citep{2025CoIR} & en & Retrieval & 13,820 & \url{huggingface.co/datasets/mteb/stackoverflow-qa} \\
ProCQA~\citep{2025procqa} & 11 & Retrieval & 485,780 & \url{github.com/jordane95/procqa} \\
Yahoo\_Answers~\citep{2015yahooanswers} & en & Retrieval & 196,645 & \url{huggingface.co/datasets/sentence-transformers/yahoo-answers} \\
GooAQ~\citep{2021gooaq} & en & Retrieval & 473,876 & \url{github.com/allenai/gooaq} \\
T2Ranking~\citep{2023t2ranking} & zh & Retrieval & 85,521 & \url{huggingface.co/datasets/sentence-transformers/t2ranking} \\
DuReader~\citep{2018dureader} & zh & Retrieval & 78,023 & \url{huggingface.co/datasets/sentence-transformers/dureader} \\
cMedQAv2~\citep{2018cmedqav2} & zh & Retrieval & 23,105 & \url{huggingface.co/datasets/sentence-transformers/cmedqa-v2} \\
Huatuo\_kgqa~\citep{2025huatuoqa} & zh & Retrieval & 53,835 & \url{huggingface.co/datasets/FreedomIntelligence/huatuo_knowledge_graph_qa} \\
Huatuo\_encqa~\citep{2025huatuoqa} & zh & Retrieval & 253,523 & \url{huggingface.co/datasets/FreedomIntelligence/huatuo_encyclopedia_qa} \\
Multi CPR Medical~\citep{2022multicprmedical} & zh & Retrieval & 62,085 & \url{github.com/Alibaba-NLP/Multi-CPR} \\
HealthCareMagic~\citep{2023healthcaremagic} & en & Retrieval & 78,626 & \url{github.com/Kent0n-Li/ChatDoctor} \\
MedicalQA\_ru~\citep{2022medicalqaru} & ru & Retrieval & 71,932 & \url{huggingface.co/datasets/blinoff/medical_qa_ru_data} \\
\midrule
\rowcolor{gray!15} \multicolumn{5}{c}{Instruction Data}\\
Aya~\citep{2024aya} & 65 & Retrieval & 126,965 & \url{huggingface.co/datasets/CohereLabs/aya_dataset} \\
MURI~\citep{2025muri} & 194 & Retrieval & 720,782 & \url{huggingface.co/datasets/akoksal/muri-it} \\
OASST2~\citep{2023oasst2} & 26 & Retrieval & 12,449 & \url{huggingface.co/datasets/OpenAssistant/oasst2} \\
MultiAlpaca~\citep{2023multialpaca} & 11 & Retrieval & 125,447 & \url{huggingface.co/datasets/DAMO-NLP-MT/multialpaca} \\
WildChat~\cite{2024wildchat} & 76 & Retrieval & 638,781 & \url{huggingface.co/datasets/allenai/WildChat-4.8M} \\
M2Lingual~\citep{2025m2lingual} & 75 & Retrieval & 158,251 & \url{huggingface.co/datasets/ServiceNow-AI/M2Lingual} \\
Natural Reasoning~\citep{2025naturalreasoning} & en & Retrieval & 845,682 & \url{huggingface.co/datasets/facebook/natural_reasoning} \\
Infinity Instruct~\citep{2025infinityinstruct} & en, zh & Retrieval & 757,439 & \url{huggingface.co/datasets/BAAI/Infinity-Instruct} \\
COIG~\citep{2025coig} & zh & Retrieval & 42,415 & \url{huggingface.co/datasets/m-a-p/COIG-CQIA} \\
Medinstruct~\citep{2023medinstruct} & en & Retrieval & 51,539 & \url{github.com/XZhang97666/AlpaCare} \\
CodeFeedbackST~\citep{2025CoIR} & 137 & Retrieval & 115,971 & \url{huggingface.co/datasets/mteb/codefeedback-st} \\
CodeFeedbackMT~\citep{2025CoIR} & python & Retrieval & 52,221 & \url{huggingface.co/datasets/mteb/codefeedback-mt} \\
OpenOrca~\citep{2023openorca} & en & Retrieval & 896,450 & \url{huggingface.co/datasets/Open-Orca/OpenOrca} \\
MEDI2~\citep{2025medi2} & en & Retrieval & 668,036 & \url{huggingface.co/datasets/GritLM/MEDI2} \\
MedicalInstruction~\citep{2023MedicalInstruction} & en & Retrieval & 75,268 & \url{huggingface.co/datasets/Mohammed-Altaf/medical-instruction-120k} \\
\midrule
\rowcolor{gray!15} \multicolumn{5}{c}{Title Matching}\\
S2ORC-Title-Abstract~\citep{2020S2ORC} & en & Retrieval & 250,000 & \url{huggingface.co/datasets/sentence-transformers/s2orc} \\
CORD 19~\citep{2020cord19} & en & Retrieval & 373,674 & \url{huggingface.co/datasets/medalpaca/medical_meadow_cord19} \\
Multi CPR ECom~\citep{2022multicprmedical} & zh & Retrieval & 90,850 & \url{github.com/Alibaba-NLP/Multi-CPR} \\
ESCI~\citep{2022esci} & en, ja, es & Retrieval & 80,468 & \url{huggingface.co/datasets/tasksource/esci} \\
CLIRMatrix~\citep{2020clirmatrix} & 137 & Retrieval & 3,275,561 & \url{github.com/ssun32/CLIRMatrix} \\
\midrule
\rowcolor{gray!15} \multicolumn{5}{c}{NLI}\\
SNLI~\citep{2015SNLI} & en & Retrieval & 54,585 & \url{huggingface.co/datasets/stanfordnlp/snli} \\
MNLI~\citep{2018MNLI} & en & Retrieval & 112,075 & \url{huggingface.co/datasets/nyu-mll/multi_nli} \\
ANLI~\citep{2020ANLI} & en & Retrieval & 18,801 & \url{huggingface.co/datasets/facebook/anli} \\
XNLI~\citep{2018XNLI} & 14 & Retrieval & 1,400,600 & \url{huggingface.co/datasets/mteb/xnli} \\
OCNLI~\citep{2020OCNLI} & zh & Retrieval & 6,616 & \url{huggingface.co/datasets/dirtycomputer/OCNLI} \\
\midrule
\rowcolor{gray!15} \multicolumn{5}{c}{Code-to-Code}\\
xCodeEval Code2Code~\citep{2024xcodeeval} & 17 & Retrieval & 37,056 & \url{huggingface.co/datasets/NTU-NLP-sg/xCodeEval} \\
xCodeEval Translation~\citep{2024xcodeeval} & 11 & Clustering & 500,000 & \url{huggingface.co/datasets/NTU-NLP-sg/xCodeEval} \\
CodeSearchNet-ccr~\citep{2025CoIR} & 6 & Retrieval & 905,195 & \url{huggingface.co/datasets/CoIR-Retrieval/CodeSearchNet-ccr} \\
    \bottomrule
    \end{tabular}
    }
\end{table}

\begin{table}[th]
    \caption{Number of samples in our collected training dataset (part 2).}
    \label{tab:data-2}
    \tiny
    \centering
    \adjustbox{width=\textwidth,center}{
    \begin{tabular}{lccrm{9cm}}
    \toprule
        \tb{Name} & \tb{Language} & \tb{Format} & \tb{Size} & \tb{URL} \\
    \midrule
\rowcolor{gray!15} \multicolumn{5}{c}{Topic Classification}\\
Arxiv Clustering P2P~\citep{2022mteb-arxiv} & en & Clustering & 83,476 & \url{huggingface.co/datasets/mteb/raw_arxiv} \\
Arxiv Clustering S2S~\citep{2022mteb-arxiv} & en & Clustering & 83,486 & \url{huggingface.co/datasets/mteb/raw_arxiv} \\
Biorxiv Clustering P2P~\citep{2022mteb-biorxiv} & en & Clustering & 57,296 & \url{huggingface.co/datasets/mteb/raw_biorxiv} \\
Biorxiv Clustering S2S~\citep{2022mteb-biorxiv} & en & Clustering & 57,296 & \url{huggingface.co/datasets/mteb/raw_biorxiv} \\
Medrxiv Clustering P2P~\citep{2022mteb-medrxiv} & en & Clustering & 18,659 & \url{huggingface.co/datasets/mteb/raw_medrxiv} \\
Medrxiv Clustering S2S~\citep{2022mteb-medrxiv} & en & Clustering & 18,659 & \url{huggingface.co/datasets/mteb/raw_medrxiv} \\
MLSUM Clustering~\citep{2020mlsum} & de, es, fr, ru & Clustering & 325,739 & \url{huggingface.co/datasets/mteb/mlsum} \\
TwentyNewsgroups~\citep{1995TwentyNewsGroups} & en & Clustering & 11,060 & \url{huggingface.co/datasets/SetFit/20_newsgroups} \\
SIB200ClusteringS2S & 205 & Clustering & 163,302 & \url{huggingface.co/datasets/mteb/sib200} \\
Reddit Clustering P2P~\citep{2021Reddit-Clustering-P2P} & en & Clustering & 80,000 & \url{huggingface.co/datasets/sentence-transformers/reddit-title-body} \\
Reddit Clustering S2S~\citep{2021Reddit-StackExchange-Clustering-S2S} & en & Clustering & 58,141 & \url{github.com/UKPLab/TWEAC-qa-agent-selection/tree/master/data/reddit/train} \\
Stack Exchange Clustering P2P~\citep{2021StackExchange-Clustering-P2P} & en & Clustering & 80,000 & \url{huggingface.co/datasets/flax-sentence-embeddings/stackexchange_title_body_jsonl} \\
Stack Exchange Clustering S2S~\citep{2021Reddit-StackExchange-Clustering-S2S} & en & Clustering & 56,731 & \url{github.com/UKPLab/TWEAC-qa-agent-selection/tree/master/data/stackexchange/train} \\
THUCNews~\citep{2016THUCNews} & zh & Clustering & 100,000 & \url{huggingface.co/datasets/SirlyDreamer/THUCNews} \\
TNews~\citep{2020CLUE} & zh & Clustering & 49,726 & \url{huggingface.co/datasets/C-MTEB/TNews-classification} \\
CSL~\citep{2022CSL} & zh & Clustering & 100,000 & \url{huggingface.co/datasets/neuclir/csl} \\
\midrule
\rowcolor{gray!15} \multicolumn{5}{c}{Summarization}\\
XSum~\citep{2018XSum} & en & Retrieval & 184,383 & \url{huggingface.co/datasets/EdinburghNLP/xsum} \\
CNN\_DM~\citep{2015CNN-DM} & en & Retrieval & 100,000 & \url{huggingface.co/datasets/abisee/cnn_dailymail} \\
MLSUM Retreival~\citep{2020mlsum} & 5 & Retrieval & 801,159 & \url{huggingface.co/datasets/mteb/mlsum} \\
Sentence Compression~\citep{2013Sentence-Compression} & en & Retrieval & 175,477 & \url{huggingface.co/datasets/sentence-transformers/sentence-compression} \\
\midrule
\rowcolor{gray!15} \multicolumn{5}{c}{Text-to-Code}\\
OCGI~\citep{2024OCGI} & python & Retrieval & 1,052,849 & \url{huggingface.co/datasets/nvidia/OpenCodeGeneticInstruct} \\
OpenCodeReasoning-2~\citep{2025OpenCodeReasoning2} & python, cpp & Retrieval & 16,632 & \url{huggingface.co/datasets/nvidia/OpenCodeReasoning-2} \\
xCodeEval NL2Code~\citep{2024xcodeeval} & 17 & Retrieval & 51,072 & \url{huggingface.co/datasets/NTU-NLP-sg/xCodeEval} \\
CosQA~\citep{2021CosQA} & python & Retrieval & 9,409 & \url{huggingface.co/datasets/mteb/cosqa} \\
SyntheticText2SQL~\citep{2024synthetic-text-to-sql} & sql & Retrieval & 99,617 & \url{huggingface.co/datasets/mteb/synthetic-text2sql} \\
\midrule
\rowcolor{gray!15} \multicolumn{5}{c}{Code-to-Text}\\
CodeSearchNet~\citep{2019CodeSearchNet} & 6 & Retrieval & 936,813 & \url{huggingface.co/datasets/CoIR-Retrieval/CodeSearchNet} \\
\midrule
\rowcolor{gray!15} \multicolumn{5}{c}{Paraphrase Detection}\\
StackExchangeDupQuestions-S2S~\citep{2021Embedding-Training-Data} & en & Retrieval & 183,559 & \url{huggingface.co/datasets/sentence-transformers/stackexchange-duplicates} \\
StackExchangeDupQuestions-P2P~\citep{2021Embedding-Training-Data} & en & Retrieval & 203,060 & \url{huggingface.co/datasets/sentence-transformers/stackexchange-duplicates} \\
QQP~\citep{2019QQP} & en & Retrieval & 243,598 & \url{gluebenchmark.com/tasks} \\
StackOverflowDupQuestions~\citep{2018StackOverflowDupQuestions} & en & Retrieval & 19,847 & \url{huggingface.co/datasets/mteb/stackoverflowdupquestions-reranking} \\
PawsX~\citep{2019pawsx} & 7 & Retrieval & 216,219 & \url{huggingface.co/datasets/google-research-datasets/paws-x} \\
\midrule
\rowcolor{gray!15} \multicolumn{5}{c}{Sentiment Analysis}\\
Amazon Polarity~\citep{2013Amazon-Reviews} & en & Classification & 100,000 & \url{huggingface.co/datasets/mteb/amazon_polarity} \\
IMDb~\citep{2011IMDB} & en & Classification & 24,904 & \url{huggingface.co/datasets/mteb/imdb} \\
Toxic Conversations~\citep{2019Toxic-Conversations} & en & Classification & 49,900 & \url{huggingface.co/datasets/mteb/toxic_conversations_50k} \\
Amazon Counterfactual~\citep{2021Amazon-Counterfactual} & en, de, ja & Classification & 14,870 & \url{huggingface.co/datasets/mteb/amazon_counterfactual} \\
Amazon Reviews~\citep{2013Amazon-Reviews} & 6 & Clustering & 600,000 & \url{huggingface.co/datasets/mteb/amazon_reviews_multi} \\
Emotion~\citep{2018Emotion} & en & Clustering & 17,944 & \url{huggingface.co/datasets/mteb/emotion} \\
Tweet Sentiment Extraction~\citep{2020tweet-sentiment-extraction} & en & Clustering & 26,732 & \url{huggingface.co/datasets/mteb/tweet_sentiment_extraction} \\
\midrule
\rowcolor{gray!15} \multicolumn{5}{c}{Intent Classification}\\
Massive Intent~\citep{2023MASSIVE} & 51 & Clustering & 661,923 & \url{huggingface.co/datasets/mteb/amazon_massive_intent} \\
MTOP Intent~\citep{2021MTOP} & 6 & Clustering & 83,922 & \url{huggingface.co/datasets/mteb/mtop_intent} \\
Banking77~\citep{2020Banking77} & en & Clustering & 9,993 & \url{huggingface.co/datasets/mteb/banking77} \\
\midrule
\rowcolor{gray!15} \multicolumn{5}{c}{Domain Classification}\\
Massive Scenario~\citep{2023MASSIVE} & 51 & Clustering & 661,923 & \url{huggingface.co/datasets/mteb/amazon_massive_scenario} \\
MTOP Domain~\citep{2021MTOP} & 6 & Clustering & 83,922 & \url{huggingface.co/datasets/mteb/mtop_domain} \\
\midrule
\rowcolor{gray!15} \multicolumn{5}{c}{Language Classification}\\
BactrianX Language Classification~\citep{2023bactrianx} & 52 & Clustering & 491,405 & \url{huggingface.co/datasets/MBZUAI/Bactrian-X} \\
\midrule
\rowcolor{gray!15} \multicolumn{5}{c}{Citation Prediction}\\
S2ORC-TItle-Citation~\citep{2020S2ORC} & en & Retrieval & 132,879 & \url{huggingface.co/datasets/sentence-transformers/s2orc} \\
S2ORC-Abstract-Citation~\citep{2020S2ORC} & en & Retrieval & 231,587 & \url{huggingface.co/datasets/sentence-transformers/s2orc} \\
SPECTER~\citep{2020SPECTER} & en & Retrieval & 24,717 & \url{huggingface.co/datasets/sentence-transformers/specter} \\
\midrule
\rowcolor{gray!15} \multicolumn{5}{c}{Linguistic Acceptability}\\
MELA~\citep{2024MELA} & 10 & Classification & 40,267 & \url{huggingface.co/datasets/Geralt-Targaryen/MELA} \\
ScaLA~\citep{2023ScaLA} & 9 & Classification & 128,471 & \url{huggingface.co/datasets/alexandrainst/scala} \\
DaLA~\citep{2025DaLA} & da & Classification & 6,508 & \url{huggingface.co/datasets/giannor/dala_large} \\
\midrule
\rowcolor{gray!15} \multicolumn{5}{c}{Claim Verification}\\
FEVER~\citep{2018FEVER} & en & Retrieval & 106,605 & \url{huggingface.co/datasets/mteb/fever} \\
SciFact~\citep{2020SciFact} & en & Retrieval & 859 & \url{huggingface.co/datasets/mteb/scifact} \\
COLIEE~\citep{2022COLIEE} & en & Retrieval & 454 & \url{www.modelscope.cn/datasets/sentence-transformers/coliee} \\
\midrule
\rowcolor{gray!15} \multicolumn{5}{c}{STS}\\
STS12~\citep{2012STS12} & en & Retrieval & 1,858 & \url{huggingface.co/datasets/mteb/sts12-sts} \\
STS22~\citep{2022STS22} & en & Retrieval & 389 & \url{huggingface.co/datasets/mteb/sts22-crosslingual-sts} \\
STSBenchmark~\citep{2021STSBenchmark} & en & Retrieval & 3,297 & \url{huggingface.co/datasets/mteb/stsbenchmark-sts} \\
STS22-Crosslingual~\citep{2022STS22} & 7 & Retrieval & 1,469 & \url{huggingface.co/datasets/mteb/sts22-crosslingual-sts} \\
BQ~\citep{2023CMTEB} & zh & Retrieval & 2,436 & \url{huggingface.co/datasets/C-MTEB/BQ} \\
    \bottomrule
    \end{tabular}
    }
\end{table}

\begin{figure}
    \centering
    \includegraphics[width=0.7\linewidth]{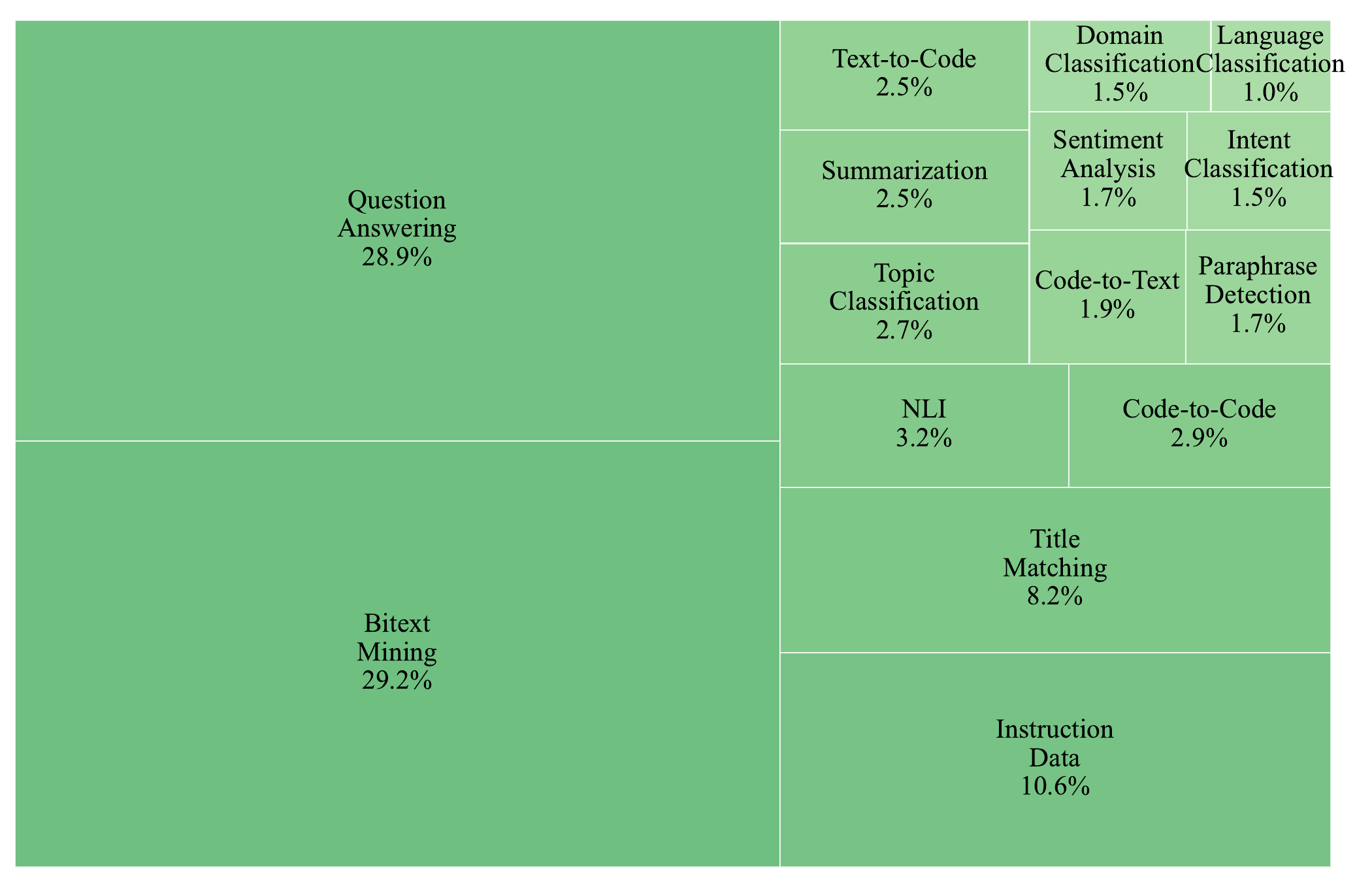}
    \caption{Task type distribution in our training data.}
    \label{fig:task-distribution}
\end{figure}

\clearpage
\section{Details on MTEB Evaluation}\label{appendix:mteb}

The Massive Text Embedding Benchmark (MTEB) is widely recognized as the de facto standard for the comprehensive evaluation of text embedding models. Originally introduced by \citet{2023MTEB}, it was vastly expanded into the Massive Multilingual Text Embedding Benchmark (MMTEB) through a large-scale, open-science collaboration~\citep{2025MMTEB}. This community-driven effort has established a rigorous and diverse evaluation framework, encompassing over 500 quality-controlled tasks that span more than 250 languages and a wide array of domains.

The significance of MTEB lies in its unprecedented scale and diversity, which addresses the critical limitations of previous benchmarks that were often constrained to a few languages (mostly English), specific domains (e.g., news), or a single task type (e.g., retrieval). To provide a holistic assessment of a model's capabilities, MTEB organizes its evaluation tasks into ten distinct categories:
\begin{itemize}
    \item Retrieval: Assesses a model's ability to find relevant documents from a large corpus for a given query.
    \item Reranking: Measures the ability to reorder a given list of candidate documents by their relevance to a query.
    \item Classification: Evaluates performance on standard text classification tasks (e.g., sentiment analysis, topic classification).
    \item Clustering: Tests how well embeddings group semantically similar documents together.
    \item Pair Classification: Involves predicting the relationship between a pair of texts (e.g., paraphrase detection, natural language inference).
    \item Semantic Textual Similarity (STS): Measures the ability to predict the degree of semantic similarity between two sentences on a continuous scale.
    \item Bitext Mining: Assesses the ability to identify translated sentence pairs from a collection of sentences in two languages.
    \item Summarization: Evaluates the semantic similarity between a model-generated summary and a reference summary.
    \item Instruction Reranking: A more challenging reranking variant where the model must follow a detailed natural language instruction to determine relevance.
    \item Multilabel Classification: A classification variant where each document can be assigned multiple labels.
\end{itemize}

The hundreds of tasks are further organized into benchmarks, which are curated subsets of tasks grouped by language, domain, or a combination of both. This includes language-specific benchmarks such as English, Chinese, and Russian; domain-specific benchmarks such as Code and Medical; and aggregated benchmarks like Multilingual, European, and Scandinavian, which test performance across a broad and diverse set of languages. This hierarchical structure allows for both a fine-grained analysis of a model's performance on a specific language or domain and a high-level view of its overall multilingual and multi-domain capabilities.

In this work, we leverage the breadth of MTEB to provide a robust and thorough evaluation of our models. We evaluate on \tb{17 benchmarks}, totaling \tb{430 unique tasks}: Multilingual, Code, Medical, English, Russian, French, German, Polish, Dutch, Indic, Persian, Chinese, Japanese, Korean, Vietnamese, European, and Scandinavian. This extensive evaluation allows for a robust and fine-grained assessment of our models' capabilities, directly supporting our claims of multilingual inclusivity and broad domain competence. The complete list of tasks used in our evaluation is detailed in Tables~\ref{tab:mteb-tasks-1}-\ref{tab:mteb-tasks-5}.

\begin{table}[ht]
    \centering
    \caption{MTEB tasks evaluated in this work: Multilingual, Code, and Medical benchmarks.}
    \label{tab:mteb-tasks-1}
    \small
    \adjustbox{width=\textwidth,center}{
    \begin{tabular}{m{1.8cm}m{15cm}}
    \toprule
        \tb{Category} & \tb{Tasks} \\
    \midrule
        \rowcolor{gray!15} \multicolumn{2}{c}{\emph{\tb{Benchmark: Multilingual}}}\\
        Bitext Mining & BornholmBitextMining, BibleNLPBitextMining, BUCC.v2, DiaBlaBitextMining, FloresBitextMining, IN22GenBitextMining, IndicGenBenchFloresBitextMining, NollySentiBitextMining, NorwegianCourtsBitextMining, NTREXBitextMining, NusaTranslationBitextMining, NusaXBitextMining, Tatoeba \\
        \midrule
        Classification & AfriSentiClassification, AmazonCounterfactualClassification, BulgarianStoreReviewSentimentClassfication, CSFDSKMovieReviewSentimentClassification, CataloniaTweetClassification, CyrillicTurkicLangClassification, CzechProductReviewSentimentClassification, DBpediaClassification, DalajClassification, EstonianValenceClassification, FilipinoShopeeReviewsClassification, FinancialPhrasebankClassification, GreekLegalCodeClassification, GujaratiNewsClassification, IndicLangClassification, IndonesianIdClickbaitClassification, IsiZuluNewsClassification, ItaCaseholdClassification, KorSarcasmClassification, KurdishSentimentClassification, MacedonianTweetSentimentClassification, MasakhaNEWSClassification, MassiveIntentClassification, MultiHateClassification, NepaliNewsClassification, NordicLangClassification, NusaParagraphEmotionClassification, NusaX-senti, OdiaNewsClassification, PAC, PoemSentimentClassification, PolEmo2.0-OUT, PunjabiNewsClassification, ScalaClassification, SentimentAnalysisHindi, SinhalaNewsClassification, SiswatiNewsClassification, SlovakMovieReviewSentimentClassification, SwahiliNewsClassification, SwissJudgementClassification, ToxicConversationsClassification, TswanaNewsClassification, TweetTopicSingleClassification \\
        \midrule
        Clustering & AlloProfClusteringS2S.v2, ArXivHierarchicalClusteringP2P, ArXivHierarchicalClusteringS2S, BigPatentClustering.v2, BiorxivClusteringP2P.v2, CLSClusteringP2P.v2, HALClusteringS2S.v2, MasakhaNEWSClusteringS2S, MedrxivClusteringP2P.v2, PlscClusteringP2P.v2, RomaniBibleClustering, SIB200ClusteringS2S, StackExchangeClustering.v2, SwednClusteringP2P, WikiCitiesClustering, WikiClusteringP2P.v2 \\
        \midrule
        Instruction Reranking & Core17InstructionRetrieval, News21InstructionRetrieval, Robust04InstructionRetrieval \\
        \midrule
        Multilabel Classification & BrazilianToxicTweetsClassification, CEDRClassification, KorHateSpeechMLClassification, MalteseNewsClassification, MultiEURLEXMultilabelClassification \\
        \midrule
        Pair Classification & ArmenianParaphrasePC, CTKFactsNLI, OpusparcusPC, PawsXPairClassification, PpcPC, RTE3, SprintDuplicateQuestions, TERRa, TwitterURLCorpus, XNLI, indonli \\
        \midrule
        Reranking & AlloprofReranking, RuBQReranking, T2Reranking, VoyageMMarcoReranking, WebLINXCandidatesReranking, WikipediaRerankingMultilingual \\
        Retrieval & AILAStatutes, ArguAna, BelebeleRetrieval, CovidRetrieval, HagridRetrieval, LEMBPasskeyRetrieval, LegalBenchCorporateLobbying, MIRACLRetrievalHardNegatives, MLQARetrieval, SCIDOCS, SpartQA, StackOverflowQA, StatcanDialogueDatasetRetrieval, TRECCOVID, TempReasonL1, TwitterHjerneRetrieval, WikipediaRetrievalMultilingual, WinoGrande \\
        \midrule
        STS & FaroeseSTS, FinParaSTS, GermanSTSBenchmark, IndicCrosslingualSTS, JSICK, SICK-R, STS12, STS13, STS14, STS15, STS17, STS22.v2, STSB, STSBenchmark, STSES, SemRel24STS \\
    \midrule
        \rowcolor{gray!15} \multicolumn{2}{c}{\emph{\tb{Benchmark: Code}}} \\
        Retrieval & AppsRetrieval, CodeEditSearchRetrieval, CodeFeedbackMT, CodeFeedbackST, CodeSearchNetCCRetrieval, CodeSearchNetRetrieval, CodeTransOceanContest, CodeTransOceanDL, CosQA, COIRCodeSearchNetRetrieval, StackOverflowQA, SyntheticText2SQL \\
    \midrule
        \rowcolor{gray!15} \multicolumn{2}{c}{\emph{\tb{Benchmark: Medical}}} \\
        Clustering & MedrxivClusteringP2P.v2, MedrxivClusteringS2S.v2 \\
        \midrule
        Retrieval & CUREv1, NFCorpus, TRECCOVID, TRECCOVID-PL, SciFact, SciFact-PL, MedicalQARetrieval, PublicHealthQA, CmedqaRetrieval \\
        \midrule
        Reranking & CMedQAv2-reranking \\
    \bottomrule
    \end{tabular}
    }
\end{table}

\begin{table}[ht]
    \centering
    \caption{MTEB tasks evaluated in this work: Russian, French, German, and Polish benchmarks.}
    \label{tab:mteb-tasks-2}
    \small
    \adjustbox{width=\textwidth,center}{
    \begin{tabular}{m{1.8cm}m{15cm}}
    \toprule
        \tb{Category} & \tb{Tasks} \\
    \midrule
        \rowcolor{gray!15} \multicolumn{2}{c}{\emph{\tb{Benchmark: Russian}}}\\
        Classification & GeoreviewClassification, HeadlineClassification, InappropriatenessClassification, KinopoiskClassification, MassiveIntentClassification, MassiveScenarioClassification, RuReviewsClassification, RuSciBenchGRNTIClassification, RuSciBenchOECDClassification \\
        \midrule
        Clustering & GeoreviewClusteringP2P, RuSciBenchGRNTIClusteringP2P, RuSciBenchOECDClusteringP2P \\
        \midrule
        Multiclass Classification & CEDRClassification, SensitiveTopicsClassification \\
        \midrule
        Pair Classification & TERRa \\
        \midrule
        Reranking & MIRACLReranking, RuBQReranking \\
        \midrule
        Retrieval & MIRACLRetrievalHardNegatives.v2, RiaNewsRetrievalHardNegatives.v2, RuBQRetrieval \\
        \midrule
        STS & RUParaPhraserSTS, STS22, RuSTSBenchmarkSTS \\
    \midrule
        \rowcolor{gray!15} \multicolumn{2}{c}{\emph{\tb{Benchmark: French}}}\\
        Classification & AmazonReviewsClassification, MasakhaNEWSClassification, MassiveIntentClassification, MassiveScenarioClassification, MTOPDomainClassification, MTOPIntentClassification \\
        \midrule
        Clustering & AlloProfClusteringP2P, AlloProfClusteringS2S, HALClusteringS2S, MasakhaNEWSClusteringP2P, MasakhaNEWSClusteringS2S, MLSUMClusteringP2P, MLSUMClusteringS2S \\
        \midrule
        Pair Classification & PawsXPairClassification \\
        \midrule
        Reranking & AlloprofReranking, SyntecReranking \\
        \midrule
        Retrieval & AlloprofRetrieval, BSARDRetrieval, MintakaRetrieval, SyntecRetrieval, XPQARetrieval \\
        \midrule
        STS & SICKFr, STSBenchmarkMultilingualSTS, STS22 \\
        \midrule
        Summarization & SummEvalFr \\
    \midrule
        \rowcolor{gray!15} \multicolumn{2}{c}{\emph{\tb{Benchmark: German}}}\\
        Classification & AmazonCounterfactualClassification, AmazonReviewsClassification, MTOPDomainClassification, MTOPIntentClassification, MassiveIntentClassification, MassiveScenarioClassification \\
        \midrule
        Clustering & BlurbsClusteringP2P, BlurbsClusteringS2S, TenKGnadClusteringP2P, TenKGnadClusteringS2S \\
        \midrule
        Pair Classification & FalseFriendsGermanEnglish, PawsXPairClassification \\
        \midrule
        Reranking & MIRACLReranking \\
        \midrule
        Retrieval & GermanQuAD-Retrieval, GermanDPR, XMarket, GerDaLIR \\
        \midrule
        STS & GermanSTSBenchmark, STS22 \\
    \midrule
        \rowcolor{gray!15} \multicolumn{2}{c}{\emph{\tb{Benchmark: Polish}}}\\
        Classification & AllegroReviews, CBD, MassiveIntentClassification, MassiveScenarioClassification, PolEmo2.0-IN, PolEmo2.0-OUT, PAC \\
        \midrule
        Clustering & EightTagsClustering, PlscClusteringS2S, PlscClusteringP2P \\
        \midrule
        Pair Classification & CDSC-E, PpcPC, PSC, SICK-E-PL \\
        \midrule
        STS & CDSC-R, SICK-R-PL, STS22 \\
    \bottomrule
    \end{tabular}
    }
\end{table}

\begin{table}[ht]
    \centering
    \caption{MTEB tasks evaluated in this work: Dutch, Indic, and Persian benchmarks.}
    \label{tab:mteb-tasks-3}
    \small
    \adjustbox{width=\textwidth,center}{
    \begin{tabular}{m{1.8cm}m{15cm}}
    \toprule
        \tb{Category} & \tb{Tasks} \\
    \midrule
        \rowcolor{gray!15} \multicolumn{2}{c}{\emph{\tb{Benchmark: Dutch}}}\\
        Classification & DutchBookReviewSentimentClassification.v2, MassiveIntentClassification, MassiveScenarioClassification, SIB200Classification, MultiHateClassification, VaccinChatNLClassification, DutchColaClassification, DutchGovernmentBiasClassification, DutchSarcasticHeadlinesClassification, DutchNewsArticlesClassification, OpenTenderClassification, IconclassClassification \\
        \midrule
        Pair Classification & SICKNLPairClassification, XLWICNLPairClassification \\
        \midrule
        Multiclass Classification & CovidDisinformationNLMultiLabelClassification, MultiEURLEXMultilabelClassification, VABBMultiLabelClassification \\
        \midrule
        Clustering & DutchNewsArticlesClusteringS2S, DutchNewsArticlesClusteringP2P, SIB200ClusteringS2S, VABBClusteringS2S, VABBClusteringP2P, OpenTenderClusteringS2S, OpenTenderClusteringP2P, IconclassClusteringS2S \\
        \midrule
        Reranking & WikipediaRerankingMultilingual \\
        \midrule
        Retrieval & ArguAna-NL.v2, SCIDOCS-NL.v2, SciFact-NL.v2, NFCorpus-NL.v2, BelebeleRetrieval, WebFAQRetrieval, DutchNewsArticlesRetrieval, bBSARDNLRetrieval, LegalQANLRetrieval, OpenTenderRetrieval, VABBRetrieval, WikipediaRetrievalMultilingual \\
        \midrule
        STS & SICK-NL-STS, STSBenchmarkMultilingualSTS \\
    \midrule
        \rowcolor{gray!15} \multicolumn{2}{c}{\emph{\tb{Benchmark: Indic}}}\\
        Bitext Mining & IN22ConvBitextMining, IN22GenBitextMining \\
        \midrule
        Clustering & SIB200ClusteringS2S \\
        \midrule
        Classification & BengaliSentimentAnalysis, GujaratiNewsClassification, HindiDiscourseClassification, SentimentAnalysisHindi, MalayalamNewsClassification, MTOPIntentClassification, MultiHateClassification, TweetSentimentClassification, NepaliNewsClassification, PunjabiNewsClassification, SanskritShlokasClassification, UrduRomanSentimentClassification \\
        \midrule
        Pair Classification & XNLI \\
        \midrule
        Retrieval & BelebeleRetrieval, XQuADRetrieval \\
        \midrule
        Reranking & WikipediaRerankingMultilingual \\
        \midrule
        STS & IndicCrosslingualSTS \\
    \midrule
        \rowcolor{gray!15} \multicolumn{2}{c}{\emph{\tb{Benchmark: Persian}}}\\
        Classification & PersianFoodSentimentClassification, SynPerChatbotConvSAClassification, SynPerChatbotConvSAToneChatbotClassification, SynPerChatbotConvSAToneUserClassification, SynPerChatbotSatisfactionLevelClassification, SynPerTextToneClassification.v3, SIDClassification.v2, DeepSentiPers.v2, PersianTextEmotion.v2, NLPTwitterAnalysisClassification.v2, DigikalamagClassification, MassiveIntentClassification, MassiveScenarioClassification, StyleClassification, PerShopDomainClassification, PerShopIntentClassification \\
        \midrule
        Clustering & BeytooteClustering, DigikalamagClustering, HamshahriClustring, NLPTwitterAnalysisClustering, SIDClustring \\
        \midrule
        Pair Classification & FarsTail, SynPerChatbotRAGFAQPC, FarsiParaphraseDetection, SynPerTextKeywordsPC, SynPerQAPC, ParsinluEntail, ParsinluQueryParaphPC \\
        \midrule
        Reranking & MIRACLReranking, WikipediaRerankingMultilingual \\
        \midrule
        Retrieval & SynPerQARetrieval, SynPerChatbotRAGFAQRetrieval, PersianWebDocumentRetrieval, WikipediaRetrievalMultilingual, MIRACLRetrievalHardNegatives, HotpotQA-FaHardNegatives, MSMARCO-FaHardNegatives, NQ-FaHardNegatives, ArguAna-Fa.v2, FiQA2018-Fa.v2, QuoraRetrieval-Fa.v2, SCIDOCS-Fa.v2, SciFact-Fa.v2, TRECCOVID-Fa.v2, FEVER-FaHardNegatives, NeuCLIR2023RetrievalHardNegatives, WebFAQRetrieval \\
        \midrule
        STS & Farsick, SynPerSTS \\
        \midrule
        Bitext Mining & SAMSumFa, SynPerChatbotSumSRetrieval, SynPerChatbotRAGSumSRetrieval \\
    \bottomrule
    \end{tabular}
    }
\end{table}

\begin{table}[ht]
    \centering
    \caption{MTEB tasks evaluated in this work: English, Scandinavian, and European benchmarks.}
    \label{tab:mteb-tasks-4}
    \small
    \adjustbox{width=\textwidth,center}{
    \begin{tabular}{m{1.8cm}m{15cm}}
    \toprule
        \tb{Category} & \tb{Tasks} \\
    \midrule
        \rowcolor{gray!15} \multicolumn{2}{c}{\emph{\tb{Benchmark: English}}} \\
        Classification & AmazonCounterfactualClassification, Banking77Classification, ImdbClassification, MTOPDomainClassification, MassiveIntentClassification, MassiveScenarioClassification, ToxicConversationsClassification, TweetSentimentExtractionClassification \\
        \midrule
        Clustering & ArXivHierarchicalClusteringP2P, ArXivHierarchicalClusteringS2S, BiorxivClusteringP2P.v2, MedrxivClusteringP2P.v2, MedrxivClusteringS2S.v2, StackExchangeClustering.v2, StackExchangeClusteringP2P.v2, TwentyNewsgroupsClustering.v2 \\
        \midrule
        Pair Classification & SprintDuplicateQuestions, TwitterSemEval2015, TwitterURLCorpus \\
        \midrule
        Reranking & AskUbuntuDupQuestions, MindSmallReranking \\
        \midrule
        Retrieval & ArguAna, CQADupstackGamingRetrieval, CQADupstackUnixRetrieval, ClimateFEVERHardNegatives, FEVERHardNegatives, FiQA2018, HotpotQAHardNegatives, SCIDOCS, TRECCOVID,  Touche2020Retrieval.v3 \\
        \midrule
        STS & BIOSSES, SICK-R, STS12, STS13, STS14, STS15, STSBenchmark, STS17, STS22.v2 \\
        \midrule
        Summarization & SummEvalSummarization.v2 \\
    \midrule
        \rowcolor{gray!15} \multicolumn{2}{c}{\emph{\tb{Benchmark: Scandinavian}}} \\
        Bitext Mining & BornholmBitextMining, NorwegianCourtsBitextMining \\
        \midrule
        Classification & AngryTweetsClassification, DanishPoliticalCommentsClassification, DalajClassification, DKHateClassification, LccSentimentClassification, MassiveIntentClassification, MassiveScenarioClassification, NordicLangClassification, NoRecClassification, NorwegianParliamentClassification, ScalaClassification, SwedishSentimentClassification, SweRecClassification \\
        \midrule
        Retrieval & DanFeverRetrieval, NorQuadRetrieval, SNLRetrieval, SwednRetrieval, SweFaqRetrieval, TV2Nordretrieval, TwitterHjerneRetrieval \\
        \midrule
        Clustering & SNLHierarchicalClusteringS2S, SNLHierarchicalClusteringP2P, SwednClusteringP2P, SwednClusteringS2S, VGHierarchicalClusteringS2S, VGHierarchicalClusteringP2P \\
    \midrule
        \rowcolor{gray!15} \multicolumn{2}{c}{\emph{\tb{Benchmark: European}}} \\
        Bitext Mining & BornholmBitextMining, BibleNLPBitextMining, BUCC.v2, DiaBlaBitextMining, FloresBitextMining, NorwegianCourtsBitextMining, NTREXBitextMining \\
        \midrule
        Classification & BulgarianStoreReviewSentimentClassfication, CzechProductReviewSentimentClassification, GreekLegalCodeClassification, DBpediaClassification, FinancialPhrasebankClassification, PoemSentimentClassification, ToxicChatClassification, ToxicConversationsClassification, EstonianValenceClassification, ItaCaseholdClassification, AmazonCounterfactualClassification, MassiveScenarioClassification, MultiHateClassification, ScalaClassification, SwissJudgementClassification, TweetSentimentClassification, CBD, PolEmo2.0-OUT, CSFDSKMovieReviewSentimentClassification, DalajClassification \\
        \midrule
        Clustering & WikiCitiesClustering, RomaniBibleClustering, BigPatentClustering.v2, BiorxivClusteringP2P.v2, AlloProfClusteringS2S.v2, HALClusteringS2S.v2, SIB200ClusteringS2S, WikiClusteringP2P.v2 \\
        \midrule
        Retrieval & StackOverflowQA, TwitterHjerneRetrieval, LegalQuAD, ArguAna, HagridRetrieval, LegalBenchCorporateLobbying, LEMBPasskeyRetrieval, SCIDOCS, SpartQA, TempReasonL1, WinoGrande, AlloprofRetrieval, BelebeleRetrieval, StatcanDialogueDatasetRetrieval, WikipediaRetrievalMultilingual \\
        \midrule
        Instruction Reranking & Core17InstructionRetrieval, News21InstructionRetrieval, Robust04InstructionRetrieval \\
        \midrule
        Multiclass Classification & MalteseNewsClassification, MultiEURLEXMultilabelClassification \\
        \midrule
        Pair Classification & CTKFactsNLI, SprintDuplicateQuestions, OpusparcusPC, RTE3, XNLI, PSC \\
        \midrule
        Reranking & WebLINXCandidatesReranking, AlloprofReranking, WikipediaRerankingMultilingual \\
        \midrule
        STS & SICK-R, STS12, STS14, STS15, STSBenchmark, FinParaSTS, STS17, SICK-R-PL, STSES \\
    \bottomrule
    \end{tabular}
    }
\end{table}

\begin{table}[ht]
    \centering
    \caption{MTEB tasks evaluated in this work: Chinese, Japanese, Korean, and Vietnamese benchmarks.}
    \label{tab:mteb-tasks-5}
    \small
    \adjustbox{width=\textwidth,center}{
    \begin{tabular}{m{1.8cm}m{15cm}}
    \toprule
        \tb{Category} & \tb{Tasks} \\
    \midrule
        \rowcolor{gray!15} \multicolumn{2}{c}{\emph{\tb{Benchmark: Chinese}}} \\
        Retrieval & T2Retrieval, MMarcoRetrieval, DuRetrieval, CovidRetrieval, CmedqaRetrieval, EcomRetrieval, MedicalRetrieval, VideoRetrieval \\
        \midrule
        Reranking & T2Reranking, MMarcoReranking, CMedQAv1-reranking, CMedQAv2-reranking \\
        \midrule
        Pair Classification & Ocnli, Cmnli \\
        \midrule
        Clustering & CLSClusteringS2S, CLSClusteringP2P, ThuNewsClusteringS2S, ThuNewsClusteringP2P \\
        \midrule
        Classification & TNews, IFlyTek, Waimai, OnlineShopping, JDReview, MultilingualSentiment, MultilingualSentiment \\
        \midrule
        STS & LCQMC, PAWSX, AFQMC, QBQTC, ATEC, BQ, STSB \\
    \midrule
        \rowcolor{gray!15} \multicolumn{2}{c}{\emph{\tb{Benchmark: Japanese}}} \\
        Clustering & LivedoorNewsClustering.v2, MewsC16JaClustering, SIB200ClusteringS2S \\
        \midrule
        Classification & AmazonReviewsClassification, AmazonCounterfactualClassification, MassiveIntentClassification, MassiveScenarioClassification, JapaneseSentimentClassification, SIB200Classification, WRIMEClassification \\
        \midrule
        Retrieval & JaqketRetrieval, MrTidyRetrieval, JaGovFaqsRetrieval, NLPJournalTitleAbsRetrieval.V2, NLPJournalTitleIntroRetrieval.V2, NLPJournalAbsIntroRetrieval.V2, NLPJournalAbsArticleRetrieval.V2, JaCWIRRetrieval, MIRACLRetrieval, MintakaRetrieval, MultiLongDocRetrieval \\
        \midrule
        Reranking & ESCIReranking, JQaRAReranking, JaCWIRReranking, MIRACLReranking, MultiLongDocReranking \\
        \midrule
        STS & JSTS, JSICK \\
    \midrule
        \rowcolor{gray!15} \multicolumn{2}{c}{\emph{\tb{Benchmark: Korean}}} \\
        Classification & KLUE-TC \\
        \midrule
        Reranking & MIRACLReranking \\
        \midrule
        Retrieval & MIRACLRetrieval, Ko-StrategyQA \\
        \midrule
        STS & KLUE-STS, KorSTS \\
    \midrule
        \rowcolor{gray!15} \multicolumn{2}{c}{\emph{\tb{Benchmark: Vietnamese}}} \\
        Retrieval & ArguAna-VN, SciFact-VN, ClimateFEVER-VN, FEVER-VN, DBPedia-VN, NQ-VN, HotpotQA-VN, MSMARCO-VN, TRECCOVID-VN, FiQA2018-VN, NFCorpus-VN, SCIDOCS-VN, Touche2020-VN, Quora-VN, CQADupstackAndroid-VN, CQADupstackGis-VN, CQADupstackMathematica-VN, CQADupstackPhysics-VN, CQADupstackProgrammers-VN, CQADupstackStats-VN, CQADupstackTex-VN, CQADupstackUnix-VN, CQADupstackWebmasters-VN, CQADupstackWordpress-VN \\
        \midrule
        Classification & Banking77VNClassification, EmotionVNClassification, AmazonCounterfactualVNClassification, MTOPDomainVNClassification, TweetSentimentExtractionVNClassification, ToxicConversationsVNClassification, ImdbVNClassification, MTOPIntentVNClassification, MassiveScenarioVNClassification, MassiveIntentVNClassification, AmazonReviewsVNClassification, AmazonPolarityVNClassification \\
        \midrule
        Pair Classification & SprintDuplicateQuestions-VN, TwitterSemEval2015-VN, TwitterURLCorpus-VN \\
        \midrule
        Clustering & TwentyNewsgroupsClustering-VN, RedditClusteringP2P-VN, StackExchangeClusteringP2P-VN, StackExchangeClustering-VN, RedditClustering-VN \\
        \midrule
        Reranking & SciDocsRR-VN, AskUbuntuDupQuestions-VN, StackOverflowDupQuestions-VN \\
        \midrule
        STS & BIOSSES-VN, SICK-R-VN, STSBenchmark-VN \\
    \bottomrule
    \end{tabular}
    }
\end{table}

\clearpage
\section{Training Details}\label{appendix:training}

\subsection{Training Hyperparameters}
We train the models with the contrastive loss
\begin{equation}
    \mathcal L = - \log\frac{e^{s(q_i,d_i^+)/\tau}}{e^{s(q_i,d_i^+)/\tau}+\sum\limits_{j=1}^{n}e^{s(q_i,d_{i,j}^-)/\tau}},
\end{equation}

where $q$ is the query, $d^+, d^-$ are positive and negative documents, and $s()$ is cosine similarity. The temperature $\tau$ is set to 0.5, and the number of hard negatives $n$ is set to $7$ (except for classification data, where $n=1$). The loss coefficients $c_{l,d'}$ in Equation~\eqref{eq:3dml_loss} are set to $c_{*,d'}=\frac{1}{\sqrt{d_{\text{model}}/d'}}$ for all layers. The models are trained with AdamW~\citep{2019AdamW}, ZeRO stage 2~\citep{2020zero}, and Flash Attention 2 enabled~\citep{2024FlashAttention2}. We set the input sequence length to 1024 during training, and the other hyperparameters are given in Table~\ref{tab:hyperparameter}.

\begin{table}[ht]
    \centering
    \caption{Training hyperparameters.}
    \label{tab:hyperparameter}
    \begin{tabular}{lrrr}
    \toprule
        Size & Learning Rate & Data Parallel & Batch Size \\
    \midrule
        0.6B & 1e-5 & 32 & 16 \\
        1.7B & 9e-6 & 32 & 16 \\
        4B & 8e-6 & 64 & 8 \\
        8B & 7e-6 & 128 & 4 \\
    \bottomrule
    \end{tabular}
\end{table}

Near the end of training, we save checkpoints at intervals of 500 steps and merge the weights of the last 5 checkpoints. Ablation experiments on a subset of four benchmarks show that this slightly but consistently improves model performance~(Table~\ref{tab:ablation-ckpt-merging}).

\begin{table}[ht]
    \centering
    \caption{Ablation results on the effectiveness of checkpoint merging.}
    \label{tab:ablation-ckpt-merging}
    \small
    \begin{tabular}{lcccc}
    \toprule
        Model & Multilingual$^{\text{(118)}}$ & English$^{\text{(37)}}$ & Code$^{\text{(10)}}$ & Medical$^{\text{(11)}}$ \\
    \midrule
        W.o. merging & 63.02 & 70.48 & 74.71 & 59.18 \\
        W. merging & 63.03 & 70.52 & 74.75 & 59.33 \\
    \bottomrule
    \end{tabular}
\end{table}

\subsection{Two-stage Training}
In the two-stage training setting, we use MMARCO, WebFAQ, CLIRMatrix, ParaCrawl, OCGI, CodeSearchNet, and CodeSearchNet-CCR for first-stage training, totaling 26.7 million samples. In the second stage, we sample at most 100 thousand queries from each data source (different language subsets within a dataset are considered to be different sources), and train the models on 8.3 million samples. As demonstrated in Table~\ref{tab:data-size-comparison}, this amount of training data is an order of magnitude smaller than the data used to train other state-of-the-art multilingual embedding models. Paired with our fully-open training data, this recipe represents a significant step forward in promoting open and reproducible embedding model research.

\begin{table}[ht]
    \centering
    \caption{Comparison of training data size (in million) among multilingual embedding models. $^*$EmbeddingGemma's training data size is estimated based on the token count reported by \citet{2025EmbeddingGemma}~(314B, 20B) and a context length of 2048. This model additionally underwent 2T tokens of encoder-decoder training from the causal model before the two-stage finetuning shown in the table.}
    \label{tab:data-size-comparison}
    \small
    \begin{tabular}{lcrr}
    \toprule
        Model & Open-data & First Stage & Second Stage \\
    \midrule
        Qwen3-Embedding & $\times$ & 150 & 12 \\
        EmbeddingGemma$^*$ & $\times$ & 153 & 10 \\
        KaLM-Embedding & $\surd$ & 100 & 5 \\
        Ours & $\surd$ & 27 & 8 \\
    \bottomrule
    \end{tabular}
\end{table}

In Table~\ref{tab:ablation-two-stage}, we present the ablation results on the impact of two-stage training conducted with the 0.6B model. We find that while the additional retrieval pre-finetuning improves performance on multilingual and English natural language performance, it comes at the cost of reduced performance on code and medical benchmarks. However, this may be related to the composition of our pre-finetuning data, and deserves in-depth investigation in the future.

\begin{table}[ht]
    \centering
    \caption{Ablation results on the effectiveness of two-stage training with the 0.6B model.}
    \label{tab:ablation-two-stage}
    \small
    \begin{tabular}{lcccc}
    \toprule
        Model & Multilingual$^{\text{(118)}}$ & English$^{\text{(37)}}$ & Code$^{\text{(10)}}$ & Medical$^{\text{(11)}}$ \\
    \midrule
        Two-stage & 63.02 & 70.48 & 74.71 & 59.18 \\
        Second-stage-only & 62.34 & 70.29 & 75.01 & 59.56 \\
    \bottomrule
    \end{tabular}
\end{table}

\clearpage
\section{Additional Results}\label{appendix:additional-results}

\subsection{3D-ML Training on Individual Languages}
In complement to the experiments in Section~\ref{sec:experiments}, we train two additional sets of models on Vietnamese (1M training samples) and Persian (350K training samples), respectively, starting from the same stage-1 0.6B checkpoint used in the main experiments. The results in Table~\ref{tab:ablation-individual-language} show that 3D-ML consistently outperforms the baseline when training on low-resource languages alone, confirming its effectiveness and wide applicability.

\begin{table}[h]
    \centering
    \caption{Comparison of baseline (140M) and 3D-ML (pruned to 140M after training) models trained individually on Vietnamese and Persian.}
    \label{tab:ablation-individual-language}
    \begin{tabular}{ccc}
    \toprule
        Model & Vietnamese & Persian \\
    \midrule
        Baseline & 40.97 & 53.88 \\
        3D-ML & 41.78 & 54.31 \\
    \bottomrule
    \end{tabular}
\end{table}

\subsection{MEL in Language Modeling}
To explore the full potential of MEL, we apply it to training causal language models. Specifically, we select Qwen2.5-0.5B-Base~\citep{2024Qwen2.5} as the backbone model to avoid benchmark contamination and saturation, and train two models on the \texttt{allenai/tulu-3-sft-olmo-2-mixture}\footnote{\url{https://huggingface.co/datasets/allenai/tulu-3-sft-olmo-2-mixture}} data: 1) baseline SFT, and 2) training with MEL. For the second mode, we factorize the embedding matrix during evaluation, reducing its parameter count to 0.4B.

For evaluation, we adopt three widely used benchmarks: GSM8K~\citep{2021gsm8k}, MMLU-Pro~\citep{2024mmlu-pro}, and WinoGrande~\citep{2019winogrande}. Interestingly, the results in Table~\ref{tab:lm-mel} indicate that performance increases despite the smaller parameter count. We hypothesize that for smaller LMs with a disproportionately large embedding layer, MEL acts as an effective regularizer and improves generalization.

\begin{table}[h]
    \centering
    \caption{Comparison of training Qwen2.5-0.5B-Base with and without MEL. All results are evaluated in 0-shot.}
    \label{tab:lm-mel}
    \begin{tabular}{cccc}
    \toprule
        Model & GSM8K & MMLU-Pro & Winogrande \\
    \midrule
        Baseline (0.5B) & 40.8 & 9.9 & 47.1 \\
        MEL (0.4B) & 43.5 & 11.1 & 50.6 \\
    \bottomrule
    \end{tabular}
\end{table}

\subsection{Efficiency Analysis}

In Table~\ref{tab:efficiency}, we report the peak GPU memory consumption and token throughput of layer-pruned models in Figure~\ref{fig:results_mll}, measured on a single A100 GPU. These results quantify the substantial, practical efficiency gains from 3D-ML. For instance, pruning the model from 28 layers to just a single layer increases throughput by over 13x (from ~43k to ~583k tokens/s), while reducing active parameters by over 70\% and peak memory by ~24\%. When viewed alongside the performance curves in Figure~\ref{fig:results_mll}, this data creates a concrete Pareto frontier, directly connecting model performance to tangible deployment metrics like latency (inversely related to throughput) and memory. However, we note that the fixed CUDA context overhead may present a confounding factor on memory measurements.

\begin{table}[h]
    \centering
    \caption{peak GPU memory consumption and token throughput of layer-pruned models in Figure~\ref{fig:results_mll}.}
    \label{tab:efficiency}
    \begin{tabular}{cccc}
    \toprule
        \# Layer & \# Parameters (M) & Peak Memory (GB) & Throughput (token/s) \\
    \midrule
        1 & 172 & 3.35 & 583831 \\
        2 & 187 & 3.63 & 318393 \\
        4 & 219 & 3.68 & 224050 \\
        8 & 282 & 3.80 & 148569 \\
        16 & 407 & 4.04 & 82832 \\
        24 & 533 & 4.27 & 58991 \\
        28 & 596 & 4.39 & 43218 \\
    \bottomrule
    \end{tabular}
\end{table}

\clearpage
\section{Limitations}
While our work demonstrates strong empirical performance and practical efficiency gains, several limitations remain:

\paragraph{Potential entanglement of method and data contributions}
Our results combine improvements from both the proposed 3D-ML framework and a newly curated multilingual dataset. Although we provide ablations to disentangle these factors on the smaller models, isolating their individual contributions remains challenging at scale. We hope our large-scale data could provide a platform to standardize comparisons of future training methods in this regard.

\paragraph{Potential dependence on base model architecture and complexity of hyperparameter choices}
Our models are built on Qwen3 causal architectures. While the 3D-ML framework is conceptually general and validated in a small-scale experiment on EuroBERT, generalization to more model architectures remains an open question.

The integration of MEL, MLL, and MRL introduces additional design choices (e.g., layer selection, rank schedules, and dimension sets). Although we provide reasonable defaults, the framework may require careful tuning for optimal performance in new settings.

\paragraph{Benchmark coverage and real-world deployment}
Despite covering 430 tasks and 17 benchmarks, our claims of inclusivity is bound by the coverage of benchmarks available in MTEB, which remains limited relative to the full long-tail of human languages. Moreover, the scores on MTEB benchmarks may not accurately reflect downstream application performance such as retrieval systems or RAG pipelines, and we suggest further validation before deploying our models in real-world applications.

\paragraph{Compute requirements for large models}
While 3D-ML improves efficiency at training and inference time, training and deploying the largest models (e.g., 8B) still requires substantial computational resources, which may limit reproducibility and accessibility.
%%%%%%%%%%%%%%%%%%%%%%%%%%%%%%%%%%%%%%%%%%%%%%%%%%%%%%%%%%%%%%%%%%%%%%%%%%%%%%%
%%%%%%%%%%%%%%%%%%%%%%%%%%%%%%%%%%%%%%%%%%%%%%%%%%%%%%%%%%%%%%%%%%%%%%%%%%%%%%%

\end{document}